\documentclass{article}

\usepackage[preprint,nonatbib]{neurips_2026}

\usepackage[utf8]{inputenc} % allow utf-8 input
\usepackage[T1]{fontenc}    % use 8-bit T1 fonts
\usepackage{hyperref}       % hyperlinks
\usepackage{url}            % simple URL typesetting
\usepackage{booktabs}       % professional-quality tables
\usepackage{tabularx}       % tables with automatic/fixed column widths
\usepackage{array}          % extended column definitions (needed for \arraybackslash, etc.)
\usepackage{amsfonts}       % blackboard math symbols
\usepackage{nicefrac}       % compact symbols for 1/2, etc.
\usepackage{microtype}      % microtypography
\usepackage{xcolor}         % colors

\usepackage{enumitem}
\setlist[itemize]{nosep, leftmargin=2.5em}
\setlist[enumerate]{nosep, leftmargin=2.5em}
\usepackage[
  backend=biber,
  style=authoryear,
  sorting=nyt,
  maxbibnames=3,
  minbibnames=1,
  maxcitenames=2,
  mincitenames=1,
  natbib=true
]{biblatex}
\usepackage{amsmath}
\usepackage{tikz}
\usetikzlibrary{positioning,arrows.meta,calc,shapes.geometric}
\usepackage{subcaption}
\usepackage{algorithm}
\usepackage[noend]{algpseudocode}
\usepackage{listings}
\usepackage{fancyvrb}
\usepackage[compact]{titlesec}

\title{The Illusion of Intervention: Your LLM-Simulated Experiment is an Observational Study}
\author{Victoria Lin\textsuperscript{\textnormal{1,2}}\thanks{Correspondence to: Victoria Lin \texttt{<victoria@stat.cmu.edu>}, Alexander D'Amour \texttt{<alexdamour@google.com>}.
}
\qquad Taedong Yun\textsuperscript{\textnormal{1}} \qquad Maja Matari\'{c}\textsuperscript{\textnormal{1}} \qquad John Canny\textsuperscript{\textnormal{1}} \\ \textbf{Arthur Gretton}\textsuperscript{1} \qquad \textbf{Alexander D'Amour}\textsuperscript{1} \\\\
\textsuperscript{1}Google DeepMind \qquad \textsuperscript{2}Carnegie Mellon University}

\begin{document}

\maketitle

\begin{abstract}
% Large language models (LLMs) are increasingly used to simulate synthetic users for applications in behavioral science, economics, policymaking, and tech, such as conducting large-scale simulated user surveys or evaluating assistive agents. 
Large language models (LLMs) show potential as simulators of human behavior, offering a scalable way to study responses to interventions.
However, because LLMs are trained largely on observational data, interventions in experiments with LLM-simulated synthetic users can induce unintended shifts in latent user attributes, causing user drift where the implicit simulated population differs across treatment conditions, potentially distorting effect estimates. We formalize the confounding or \textit{selection bias} that can arise due to user drift and show how intervention-dependent shifts can inflate or attenuate observed differences in user responses under intervention.
% that correspond not to the causal treatment effect alone but rather the causal treatment effect \textit{plus} a selection-bias term. 
To diagnose confounding, we propose using negative control outcomes---attributes that should remain invariant under intervention---to identify distribution shifts across intervention conditions, providing evidence of user drift. To mitigate drift, we study adjusting the persona specification by eliciting additional confounders, finding that targeted, setting-relevant confounders can substantially reduce bias across survey-style and multi-turn agent evaluations. 
% \cmt{TY: We should probably add a sentence or two about the datasets and results. The character limit is 2000 and we've used 1273 so far, so we have space.}
\end{abstract}

\setcounter{footnote}{0}

\section{Introduction}
Large language models (LLMs) are emerging as powerful tools for simulating human behavior. Traditionally, a multitude of scientific applications rely on human feedback. Large-scale surveys underpin many social science and public policy studies, and modern machine learning systems are trained and evaluated on human response data. Because collecting such feedback at scale is a major bottleneck, a popular alternative is to use LLMs to simulate users \citep{manning2024automatedsocialsciencelanguage, Argyle_Busby_Fulda_Gubler_Rytting_Wingate_2023}, often by prompting them to embody a persona \citep{Gu_Chandrasegaran_Lloyd_2025, frohling-etal-2025-personas}. These users can then be used in lieu of human respondents to conduct \textit{synthetic experiments} that explore how different populations might respond to new policies, products, or conversational interactions---for instance, evaluating whether an \textit{intervention} such as an update to an AI agent yields more favorable user outcomes \citep{li2025llm, kaiser2025simulating}.

In this setting, synthetic users\footnote{We use ``synthetic user'' to mean an LLM initialized to simulate a user with a short persona prompt (e.g., you are a 30-year-old man). This defines a distribution over users, not a single user, and we operate at this level in our formal treatment.} are instantiated by sampling personas from a target user distribution. 
% \john{Since the $L$s are partial descriptions, and we're using an $L,A,X,Y$ generator, $X$ remains a non-point distribution over users throughout the interaction with the LLM. We never really instantiate a single user the way the original causal model does.}
Each synthetic user is then assigned to interact with each version of an AI agent and provide feedback. Ideally, the synthetic users' responses and derived test statistics serve as valid counterfactuals, reflecting how real users would respond in the same experiment and enabling iterative improvement of the agent \citep{bai2022constitutionalaiharmlessnessai, lee2024rlaif, castricato-etal-2025-persona, yun-etal-2025-sleepless}.

\begin{figure}
    \centering
    \includegraphics[width=0.65\textwidth]{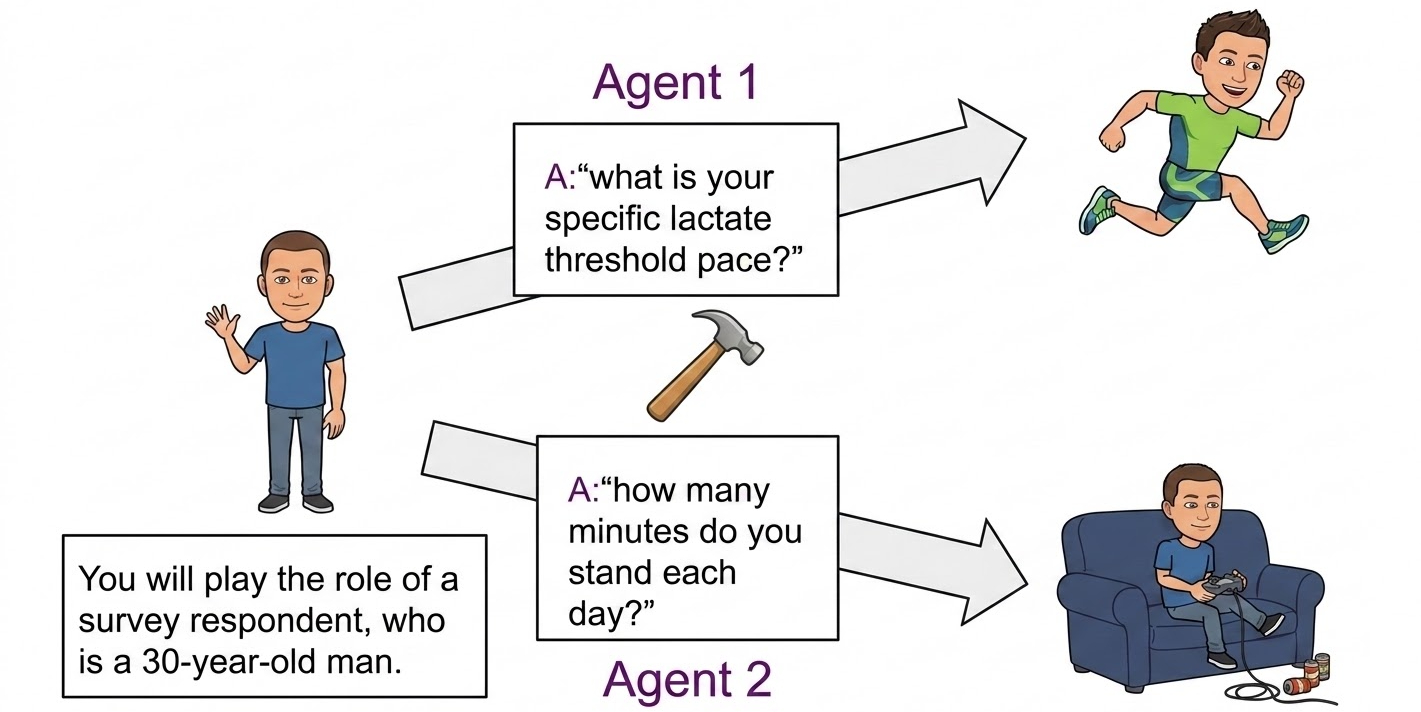}
    \medskip
    \caption{In LLM-simulated experiments, two synthetic users who are instantiated with identical personas can drift to become distinct users in response to different intervention conditions.}
    \label{fig:overview}
\end{figure}

In this paper, we argue that this vision is overly optimistic 
% \john{for LLMs using short prompts (aside: so far we havent seen these effects on PT models with long backstories)}
due to \textit{user drift} in LLM simulations. While synthetic experiments are designed to mimic randomized controlled trials, LLMs are trained to operate via abductive reasoning: given underspecified prompts, they infer as much unspecified context as they are able \citep{wright-etal-2024-llm, lee2026llmsinferpoliticalalignment}. Consider a synthetic experiment evaluating 
% two movie recommender agents, where the user is initialized as a 30-year-old man. Agent 1 is prompted to suggest lesser-known films, while Agent 2 suggests mainstream films. To generate a coherent conversation, the LLM interacting with Agent 1 might implicitly adopt the latent traits of an indie film enthusiast, whereas the LLM interacting with Agent 2 might simulate a casual moviegoer.
two health coaching agents, where the user is initialized as a 30-year-old man (Figure \ref{fig:overview}). Agent 1 is prompted to discuss the user's exercise habits and physical condition in a more technical manner, while Agent 2 discusses physical activity in more general terms. To generate a coherent conversation, the LLM interacting with Agent 1 might implicitly adopt the latent traits of a serious athlete, while the LLM interacting with Agent 2 might simulate a less active individual.
% \john{Not sure I would characterize this as ``drift''. The posterior distribution over possible personas is shrinking as more dialog is generated by and to the model, and the regions it shrinks to are influenced by what its interlocutor says. But its not as though there is a point user posterior that is moving...}

% two LLMs instantiated with the same persona---when exposed to different AI agents---can drift in persona attributes that (1) would remain fixed in human subjects across agent interactions, and (2) are relevant to the primary outcomes of the experiments.

As a result, two synthetic users who are instantiated with the same persona at the beginning of an experiment can effectively become distinct users by the time their outcomes are measured. 
% \john{No, there was a distribution of potential users at the beginning that included the final distributions, and that shrank to those different distributions as a result of dialog.}
% This drift can obfuscate the effect of the intervention and lead to misleading conclusions: better simulated outcomes for one agent may reflect either a genuine improvement for a fixed user population (a causal effect) or a shift in the simulated user population across agent versions.
Because this drift 
% \john{dialog-driven variation?} 
can occur over persona attributes that (1) would remain fixed in human subjects and (2) are relevant to the primary outcomes of the experiment, it can obfuscate the actual intervention effect and give rise to misleading results. If the synthetic user reports higher satisfaction with Agent 2, it is unclear whether Agent 2 is actually more helpful or if the simulated user has simply drifted into a population that is easier to please. 
% \john{Thus the final subject population can not be guaranteed to be representative of an intended target population as in a human experiment.}
% For instance, if a change to an AI agent yields better simulated outcomes, it is not clear whether this reflects an improvement in how a fixed user population would experience the agent (i.e., the causal effect of the update), or that the LLM is simulating a distinct user population that reacts more favorably to agent interactions. While this behavior is undesirable in synthetic experiments, it results from LLMs operating as intended: given underspecified prompts, they infer as much unspecified context as they are able (i.e., \textit{abductive reasoning}). It is important to develop strategies to detect and mitigate user drift of this kind.

% \textcolor{red}{[TODO: Example?]}

To diagnose and mitigate this issue, our work makes three contributions:
\begin{enumerate} 
\item We formalize the bias induced by user drift through the lens of \textit{confounding} in causal inference. We argue that because LLMs are trained largely on observational data, synthetic user experiments are more similar to observational studies than randomized experiments. As a result, differences in conditional user populations given the intervention---a form of \textit{selection bias}---can confound the estimated treatment effect. 

\item We introduce an empirical diagnostic for LLM user drift 
% method to detect drift in LLM-simulated users empirically 
using \textit{negative control outcomes} \citep{lipsitch2010negative}: variables pertaining to the user that should remain invariant under intervention, such as the user's race. Differences in negative control outcome distributions across intervention conditions indicate that the latent attributes of the simulated user---which are causally and temporally upstream of the intervention and would remain fixed in a real experiment---do change in response to the intervention. 
We provide evidence of confounding in simulated users across multiple major families of open-source and proprietary LLMs.

\item We show that user drift and selection bias can be mitigated in simulated users via \textit{confounder adjustment} during user persona specification.
% , in a procedure philosophically akin to adjusting for confounders in traditional observational causal inference settings. 
We present evidence that 
% \ag{meaning?} 
selection of confounders that are explicitly relevant to the treatment and outcome of the synthetic experiment helps to reduce selection bias, while adjustment over attributes that are less directly relevant yields inconsistent benefits and can sometimes inadvertently exacerbate selection bias. 
% \john{I dont think a typical observer would see that in the figures. For OpinionQA, most of the TVD reduction is caused by the demographic confounders. For the other two, there is substantial TVD reduction from the demographic confounders (albeit with a fair amount of superimposed noise). During the targeted confounder phase, only one or two of the trajectories show TVD reduction, while it increases for others. Ideally we would propose a hypothesis to test this, and make sure we have confirmation of it. The plots now have a lot of variance in them...}
\end{enumerate}
% \ag{more specifically?}. 

% We make this observation precise by showing that the observed difference in a synthetic experiment decomposes into the target causal effect and a selection bias.
% \textcolor{red}{[TODO: State up front here that this is actually LLMs functioning as intended/trained - they make inferences about unspecified context given an under/partially specified prompt.]} 
% Second, we introduce an empirical method to detect drift in LLM-simulated users via \textit{negative control outcomes}: user variables that should be invariant to the intervention (e.g., fixed demographic attributes such as race). Differences in negative control outcome distributions across intervention conditions indicate that the latent background information of the simulated user---which is causally and temporally upstream of the intervention and would remain fixed in a real experiment---changes in response to the intervention. 

Our findings suggest that evaluations with synthetic users need to be treated with care. This work highlights how the generative interface of LLMs can be misleading: while an evaluation may be designed to mimic a randomized controlled trial, the LLM-simulated users employed in the trial can introduce confounding bias.

% \ag{I suggest putting  a paper outline here: "In Section 1, ...". Gemini can probably write it.}

% \renewcommand{\john}[1]{\textcolor{orange}{[John: #1]}}

% \vspace{-0.25em}
\section{Related work}
\label{sec:related-work}
% LLMs are increasingly used for user simulation.
% Persona prompting is a particularly popular strategy.
% Studies have shown varying success and improvement in persona stability?, making this approach increasingly technically feasible.
% Most studies have focused on external validity, such as quality of simulation, and fidelity of generated rating or opinion patterns within particular persona groupings.
% Our work adds to this area by focusing on internal validity of interventional comparisons. This is a distinct question and complements work on external validity.

\textbf{LLM user simulation.} \; %LLMs are increasingly used to simulate user behavior. 
LLMs are increasingly used to simulate user behavior, and persona prompting has become a particularly popular strategy for generating diverse user perspectives. Simulation approaches include both using short persona prompts containing demographic attributes \citep{park2023generative, hu-collier-2024-quantifying, kaiser2025simulating, frohling-etal-2025-personas} and longer, life-story-based persona prompting \citep{moon2024virtualpersonaslanguagemodels, park2026llmagentsgroundedselfreports},
which provides benefits such as ``deep binding'' \citep{kang2025deep} and more accurate modeling of population variance.
Because conversational interfaces are the standard for synthetic user interactions, the majority of the simulation literature evaluates instruction-tuned LLMs. Prior works analyzing these specific simulators have identified distinct challenges, such as mode collapse, sycophancy, and persona instabilities \citep{anthis2025social,li2025llm, Tosato_Helbling_Mantilla-Ramos_Hegazy_Tosato_Lemay_Rish_Dumas_2026}. Some evidence suggests these challenges are less pronounced in simulations with base pre-trained models \citep{lyman2025balancing}, though they remain subject to artifacts such as prompt sensitivity \citep{sclar2024quantifyinglanguagemodelssensitivity}.

Much of the work on evaluating LLM-simulated users has centered on their \textit{external validity}: how closely the generated ratings or opinion patterns of simulated users match those of their real human counterparts. Recent works develop benchmarks for external validity \citep{dou-etal-2025-simulatorarena}, identify failures \citep{Aher2022, li2025llm, gao2025cautionusingllmshuman, Neumann_De-Arteaga_Fazelpour_2026}, and propose strategies to better align synthetic users with human ones \citep{kang2025deep, moon2026identitycooperationframingeffects, dong-etal-2024-llm, guerdan2026doublyrobust}. Our work is complementary to this line of research: it focuses on the \textit{internal validity} of the synthetic user study under intervention. This is a distinct question that specifically examines whether the difference between experimental conditions can be interpreted causally, rather than whether the results of the experiment can be generalized.

% A relevant quantity and the target of many benchmarks has been to assess how closely findings or statistics from studies with LLM users match those of real human users---that is, the \textit{external validity} of LLM users. Recent works diagnose failures in external validity

% \begin{itemize}
%     \item Persona instability: \citet{Tosato_Helbling_Mantilla-Ramos_Hegazy_Tosato_Lemay_Rish_Dumas_2026} 
%     \item Evaluating/diagnosing failures in external validity (i.e., alignment with humans): \citet{Aher2022, li2025llm, gao2025cautionusingllmshuman, Neumann_De-Arteaga_Fazelpour_2026}. Benchmarks: \citet{dou-etal-2025-simulatorarena}
%     \item Improving external validity: \citet{kang2025deep, moon2026identitycooperationframingeffects} (backstories/deep binding), \citet{dong-etal-2024-llm} (verbalized confidence), \citet{guerdan2026doublyrobust} (DR estimator)
% \end{itemize}

\textbf{Confounding in LLMs.} \; The most similar prior work on this topic is from \citet{Gui_2023}, who noted that GPT-4o-mini 
\citep{openai2024gpt4ocard} 
exhibits empirical behaviors consistent with confounding in simulated blinded microeconomic experiments. 
% In their work, text-based treatments correlated with shifts in unspecified variables that would have remained constant in a real experimental setting. 
Our work builds on this empirical study by introducing a formalism for confounding in LLM user simulations. We further propose distinct, unified approaches to diagnose and adjust for confounding bias in a more general experimental setting.

\section{Problem statement}
In real and synthetic experiments, the goal is to estimate the causal effect of an intervention on a user's response.
With human subjects, the gold standard for establishing causality is a randomized trial, because the randomized treatment assignment ensures that background factors that might inform user responses are balanced across groups that receive different interventions. This isolates the effect of the intervention on the outcome.
Synthetic experiments seek to replicate this property with synthetic users: identically initialized LLMs are exposed to different interventions, with the expectation that this yields a controlled comparison.
In this section, we argue that what LLMs actually generate in a synthetic experiment more closely resembles data from an observational study.

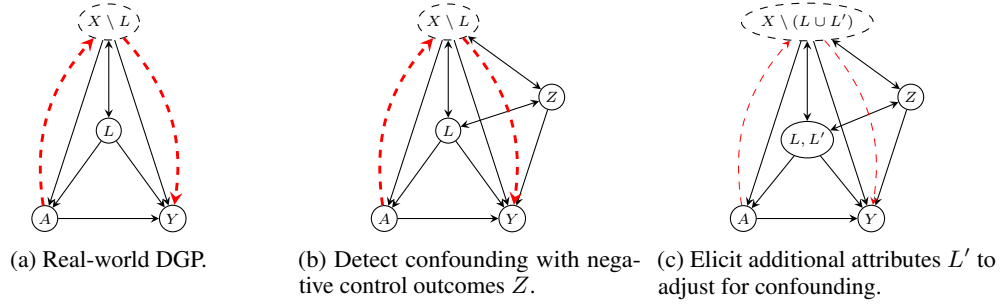
\begin{figure*}[t]
    \centering

    % (a) Base graph
    \begin{subfigure}[t]{0.32\linewidth}
        \centering
        \begin{tikzpicture}[>=stealth, scale=0.8, transform shape, every node/.style={font=\scriptsize}]
            % Keep a pentagram-ish layout by placing nodes on (unused) pentagon vertices.
            \def\R{1.8cm}
            \node[draw, ellipse, dashed, inner xsep=0pt, inner ysep=2.5pt] (XmL) at (90:\R) {$X\setminus L$};
            \node[draw, circle, inner sep=1.6pt] (A) at (234:\R) {$A$};
            \node[draw, circle, inner sep=1.6pt] (Y) at (306:\R) {$Y$};
            \node[draw, circle, inner sep=1.6pt] (L) at (0,0) {$L$};

            % Main edges
            \draw[<->] (XmL) -- (L);
            \draw[->] (XmL) -- (A);
            \draw[->] (XmL) -- (Y);
            \draw[->] (A) -- (Y);
            \draw[->] (L) -- (A);
            \draw[->] (L) -- (Y);

            % Treatment-induced persona shift / selection artifact
            \draw[->, red, dashed, line width=1.1pt] (A) to[bend left=25] (XmL);
            \draw[->, red, dashed, line width=1.1pt] (XmL) to[bend left=25] (Y);
        \end{tikzpicture}
        \caption{Real-world DGP.}
        \label{fig:dag:conceptual}
    \end{subfigure}\hfill
    % (b) Add negative control $Z$
    \begin{subfigure}[t]{0.32\linewidth}
        \centering
        \begin{tikzpicture}[>=stealth, scale=0.8, transform shape, every node/.style={font=\scriptsize}]
            \def\R{1.8cm}
            \node[draw, ellipse, dashed, inner xsep=0pt, inner ysep=2.5pt] (XmL) at (90:\R) {$X\setminus L$};
            \node[draw, circle, inner sep=1.6pt] (Z) at (18:\R) {$Z$};
            \node[draw, circle, inner sep=1.6pt] (A) at (234:\R) {$A$};
            \node[draw, circle, inner sep=1.6pt] (Y) at (306:\R) {$Y$};
            \node[draw, circle, inner sep=1.6pt] (L) at (0,0) {$L$};

            % Main edges
            \draw[<->] (XmL) -- (L);
            \draw[->] (XmL) -- (A);
            \draw[->] (XmL) -- (Y);
            \draw[->] (A) -- (Y);
            \draw[->] (L) -- (A);
            \draw[->] (L) -- (Y);

            % Negative-control structure
            \draw[<->] (XmL) -- (Z);
            \draw[<->] (L) -- (Z);
            \draw[->] (Z) -- (Y);

            % Treatment-induced persona shift / selection artifact
            \draw[->, red, dashed, line width=1.1pt] (A) to[bend left=25] (XmL);
            \draw[->, red, dashed, line width=1.1pt] (XmL) to[bend left=25] (Y);
        \end{tikzpicture}
        \caption{Detect confounding with negative control outcomes $Z$.}
        \label{fig:dag:negative-control}
    \end{subfigure}\hfill
    % (c) Rename $X\setminus L$ and $L$ to set-valued versions; dashed red edges are not bold
    \begin{subfigure}[t]{0.32\linewidth}
        \centering
        \begin{tikzpicture}[>=stealth, scale=0.8, transform shape, every node/.style={font=\scriptsize}]
            \def\R{1.8cm}
            % Use ellipses for long set-valued labels (avoid the "vertical ends" of rounded rectangles).
            \node[draw, ellipse, dashed, inner xsep=0pt, inner ysep=2.5pt] (XmLLp) at (90:\R) {$X\setminus (L \cup L')$};
            \node[draw, circle, inner sep=1.6pt] (Z) at (18:\R) {$Z$};
            \node[draw, circle, inner sep=1.6pt] (A) at (234:\R) {$A$};
            \node[draw, circle, inner sep=1.6pt] (Y) at (306:\R) {$Y$};
            % Slightly narrower center node so surrounding edges can be straight.
            \node[draw, ellipse, inner xsep=0pt, inner ysep=2.5pt] (LLp) at (0,-1.5mm) {$L, L'$};

            % Main edges
            \draw[<->] (XmLLp) -- (LLp);
            \draw[->] (XmLLp) -- (A);
            \draw[->] (XmLLp) -- (Y);
            \draw[->] (A) -- (Y);
            \draw[->] (LLp) -- (A);
            \draw[->] (LLp) -- (Y);

            % Negative-control structure (also present in (c))
            \draw[<->] (XmLLp) -- (Z);
            \draw[<->] (LLp) -- (Z);
            \draw[->] (Z) -- (Y);

            % Treatment-induced persona shift / selection artifact (not bold)
            \draw[->, red, dashed, line width=0.3pt] (A) to[bend left=25] (XmLLp);
            \draw[->, red, dashed, line width=0.3pt] (XmLLp) to[bend left=25] (Y);
        \end{tikzpicture}
        \caption{Elicit additional attributes $L'$ to adjust for confounding.}
        \label{fig:dag:adjust}
    \end{subfigure}

    \caption{DAGs for (a) the real-world observational data-generating process, (b) detecting selection bias via negative controls $Z$, and (c) reducing selection bias by iteratively eliciting additional attributes $L'$ from $X \setminus L$. Red dashed arrows indicate open backdoor paths that drive the mechanism of user drift and selection bias in simulated user populations. 
    % \ag{add explicit terminology "back door path" for red arrows to figure caption}
    }
    \label{fig:dgp}
\end{figure*}

\textbf{Notation and task.} \;
Let $A \in \{0, 1\}$ denote a treatment or intervention on a user, and let $Y$ denote the primary outcome (the user's response). Let the full underlying user persona be represented by a set of attributes $X$. Let $L$ be the subset of those attributes in $X$ that are observed, and let $X\setminus L$ be the remaining latent attributes not included in $L$. The task is then to estimate how intervening on $A$ changes the distribution of $Y$ \textit{in a real user population.} In an agent evaluation setting, for instance, $A$ might be a dialogue with Agent 1 
% \cmt{TY: maybe agent \#1 and 2 (or alpha, beta, ...), given A is also used for actions} 
($A=0$) or Agent 2 ($A=1$), and $Y$ might be a user-reported measure of conversation satisfaction or quality. Variables are illustrated in Figure~\ref{fig:dag:conceptual}. 
% \john{In our case, don't we just need a directed arrow from $X$ to $L$ (not a mutual cause one)?}

\textbf{Ideal randomized setting.} \; 
% \ag{Confusing: looks like you're starting an "experiments" section.}
We can now express a core distinction between real and synthetic experiments.
Following the potential outcomes framework \citep{rubin1974estimating}, let $Y(a)$ denote the potential (or counterfactual) outcome that would be observed under intervention $A=a$.
We aim to estimate the \textit{average treatment effect} (ATE)
\begin{equation}
    \tau^{\text{ATE}} = \mathbb{E}[Y(1) - Y(0)] = \mathbb{E}[\mathbb{E}[Y(1) - Y(0) \mid L]].
    \label{eq:cate}
\end{equation}

We also make two standard causal assumptions: (1) \textit{Consistency}: $Y = AY(1) + (1-A)Y(0)$ and (2) \textit{Latent unconfoundedness}: Given $X$, $A \perp Y(0),Y(1)$.
% \begin{enumerate}
% \item \textit{Consistency}: $Y = AY(1) + (1-A)Y(0)$
% \item \textit{Latent unconfoundedness}: Given $X$, $A \perp Y(0),Y(1)$
% \end{enumerate}
Under these assumptions, we can represent
\begin{equation*}
\mathbb E[Y(a) \mid L] = \int_\mathcal{X} \underbrace{\mathbb E[Y \mid A=a, X \mid L]}_{\text{outcome-generating process}} \underbrace{P(X \mid L)}_{\text{fixed population}} dX.
\end{equation*}
This is the average outcome across the fixed marginal population of personas $X$ for both $Y(1)$ and $Y(0)$, which is the quantity measured in real randomized trials. $\tau^{\text{ATE}}$ is then identified from the data, as $\mathbb{E}[Y(a) \mid L]=E[Y \mid A=a, L]$.

\textbf{Synthetic setting.} \; 
In a synthetic experiment, instead of testing 
% how the intervention $A$ affects outcomes $Y$ for 
a real population of users, we prompt an LLM to embody many different personas by specifying user attributes $L$. For each persona, we simulate interactions under multiple conditions of $A$ (e.g., Agent 1 
% \cmt{TY: same as above, it's a bit confusing to use A again} 
vs. Agent 2) and outcomes $Y$.
% This allows us to observe how identically-initialized LLMs generate different outcomes in response to different interventions. 
Superficially, this resembles a controlled comparison in which different interventions are presented to an identical user. This setup is enabled by how the LLM models its generative distribution. While the real-world data-generating process follows the causal ordering $X, L, A, Y$, the LLM factorizes the joint distribution in the sequence of its prompt context: $L, A, X, Y$. As a generative procedure, this ordering is convenient: it lets us condition on observed values of $L$ and assign both interventions $A$ to every persona, yielding a minimum-variance estimate of the apparent treatment effect. 

However, LLMs are trained on observational data in which a user's true persona $X$ can influence both their likelihood of experiencing a treatment $A$ (e.g., self-selection into interacting with Agent 1 vs. 2) and their eventual response $Y$. Consequently, when we introduce the treatment $A=a$ to the LLM's context, which has also previously been instantiated with $L$, the model generates its response---and thus the outcome---from an \textit{observed} conditional distribution rather than an interventional one:
% \begin{equation}
% \begin{split}
% &P(Y \mid A=a, L) = \\
% &\, \int_X \underbrace{P(Y \mid A=a, X, L)}_{\text{outcome-generating process}}
%    \underbrace{P(X \mid A=a, L)}_{\text{treatment-conditional population}} dX
% \end{split}
% \label{eq:observed}
% \end{equation}
% This yields the \textit{observed effect}:
% \begin{equation*}
%     \tau=\mathbb{E}[Y \mid A=1,L]-\mathbb{E}[Y \mid A=0,L]
%     \label{eq:naive-cate}
% \end{equation*}
{%
\setlength{\abovedisplayskip}{5pt}%
\setlength{\belowdisplayskip}{5pt}%
\setlength{\abovedisplayshortskip}{5pt}%
\setlength{\belowdisplayshortskip}{5pt}%
\begin{equation}
\mathbb E[Y \mid A=a, L] =  \int_\mathcal{X} \underbrace{\mathbb{E}[Y \mid A=a, X, L]}_{\text{outcome-generating process}}
   \underbrace{P(X \mid A=a, L)}_{\text{treatment-conditional population}} dX
\label{eq:observed}
\end{equation}
}%
% \vspace{-0.2em}
This yields the \textit{observed effect}: 
{%
\setlength{\abovedisplayskip}{6pt}%
\setlength{\belowdisplayskip}{6pt}%
\setlength{\abovedisplayshortskip}{6pt}%
\setlength{\belowdisplayshortskip}{6pt}%
\begin{equation*}
    \tau^{\text{obs}}=\mathbb{E}[Y \mid A=1] - \mathbb{E}[Y \mid A=0] = \mathbb{E}[\mathbb{E}[Y \mid A=1,L]-\mathbb{E}[Y \mid A=0,L]]
    \label{eq:naive-cate}
\end{equation*}
}%
% the intervention $A$ appears in the context of an LLM, it generates its response (and thus the outcome) from a conditional distribution of text given its context, which includes its persona specification and the agent action: $P(Y \mid A=a, L).$
% \textbf{Notation.}
% \begin{itemize}
%     \item $X$: Unobserved ``true persona'' of the user
%     \item $L$: User attributes specified in prompt
%     % \item $Z$: Additional user attributes not specified in prompt
%     \item $A$: Intervention on user, which here is some kind of utterance or conversation that the user interacts with
%     \item $Y$: User's outcome
% \end{itemize}
% However, an LLM is trained on observed data, where a user's true persona $X$ dictates both their likelihood of experiencing a treatment $A$ (e.g., self-selecting into reading a conservatively-biased or liberally-biased statement) and their eventual outcome $Y$ (Figure \ref{fig:dgp}). Therefore, the LLM can produce only the \textit{observed} distribution
% $$P(Y \mid A=a, L) = \int_X P(Y \mid A=a, X, L)P(X \mid A=a, L)\; dX$$
% This leads to the g contrast
% \begin{equation}
%     \tau=P(Y \mid A=1,L)-P(Y \mid A=0,L)
% \end{equation}
% which equals \eqref{eq:cate} only if $P(X \mid A=1, L) = P(X \mid A=0, L)$.
What \eqref{eq:observed} suggests is that the distribution of user attributes $X$---and in particular the distribution of unobserved  user attributes $X \setminus L$, as $L$ is specified---can differ based on the value of $A$. 
% \john{But don't we similarly expect $X$ to be affected by different wordings of the traits $L$ based on the papers we cited? I have to admit I don't know what $X\setminus L$ means in this context... $L$ induces a distribution over $X$ rather than defining a subset of parameters??}
% We contrast this with the ideal potential outcome distribution, where the user distribution remains fixed:
% \begin{equation*}
% P(Y(a) \mid L) = \int_X \underbrace{P(Y \mid A=a, X, L)}_{\text{outcome-generating process}}
%    \underbrace{P(X \mid L)}_{\text{fixed population}} dX
% \label{eq:potential_outcome}
% \end{equation*} 
This means that in a synthetic experiment, rather than comparing identical users' responses to different treatments, the initial identical users may \textit{drift} to become distinct individuals with differing latent user attributes. 
While this occurs because the model generates responses by inferring a coherent persona from the entire context, this mechanism operates against the arrow of time, as a later intervention can influence what is meant to be the user's initial state.
% the model's inferences occur against the arrow of time, since post-intervention it "sets" user attributes that should be determined prior to the intervention.

Formally, if there is no confounding, as in a randomized experiment, then $P(X \mid A=1, L)=P(X \mid A=0, L)$, and $\tau^{\text{obs}}=\tau^{\text{ATE}}$. If $P(X \mid A=1,L)\neq P(X \mid A=0, L)$---that is, if the distribution of latent user attributes differs over $A$---then $P(Y \mid A=1, L)$ and $P(Y \mid A=0, L)$ 
effectively compare outcomes generated under different user populations, creating an internal validity problem that biases the observed effect.
% where the conditional outcome distributions are no longer comparable due to \textit{selection bias} in the user populations associated with each intervention condition. 
% If there is no confounding, as in a randomized experiment, then $P(X \mid A=1, L)=P(X \mid A=0, L)$, and $\tau=\tau^*$. 
We show an explicit decomposition for this bias in Appendix \ref{sec:ate_decomposition}.
% : the intervention conditions are no longer comparable because they are evaluated over different latent user populations.
% differ not only in $A$ but also in the induced distribution over $X$.
% In turn, the LLM will produce outcome distributions corresponding to distinct treatment-conditional populations.

% Crucially, if $P(X \mid A=1,L)\neq P(X \mid A=0, L)$---that is, if the distribution of unspecified latent user attributes differs based on the value of $A$---then the two conditions $P(Y \mid A=1, L)$ and $P(Y \mid A=0, L)$ effectively compare outcomes generated under different latent user populations. This creates an internal validity problem: the intervention conditions are no longer comparable because they differ not only in $A$ but also in the induced distribution over $X$.

% In a randomized experiment (stratified on $L$), random assignment implies $A \perp X \mid L$, and therefore $P(X \mid A=1, L)=P(X \mid A=0, L)$. Under standard causal assumptions (consistency, exchangeability, positivity), this yields $P(Y(a) \mid L)=P(Y \mid A=a, L)$.

Figure~\ref{fig:dag:conceptual} depicts this issue as an open \emph{backdoor path} from $A$ to $Y$ through latent user attributes. In the observational data that LLMs learn from, the unspecified attributes $X\setminus L$ influence both treatment exposure and outcomes, yielding the confounding path $A \leftarrow (X\setminus L) \rightarrow Y$. In an ideal randomized experiment, random assignment (within strata of $L$) blocks these backdoor paths by enforcing $A \perp (X\setminus L) \mid L$. In contrast, in LLM-simulated users instantiated with attributes $L$, interactions with different interventions $A$ can themselves shift which unspecified latent attributes are inferred. Backdoor confounding in the training data-generating process induces shifts in $X \setminus L$ at simulation time, creating a post-treatment path from $A$ to $Y$.
% , opening backdoor paths from $A$ to $Y$ by changing the distribution of $X \setminus L$ across treatment conditions.
% and presented with intervention $A$, the model generates $Y$ by inferring a coherent latent persona from the entire context. This means that the later intervention can---against the arrow of time---determine the user's unspecified latent attributes $X \setminus L$ (the red dashed edges). This effectively re-opens backdoor paths by changing the distribution of $X\setminus L$ across treatment conditions. 

Our goal is therefore to detect these backdoor-induced biases by using negative control outcomes $Z$ (Figure~\ref{fig:dag:negative-control}) and to reduce them by eliciting additional attributes $L'$ from $X\setminus L$ (Figure~\ref{fig:dag:adjust}). This ensures that comparisons across $A$ are made over more similar latent user populations.
% \john{Does the above work if we replace $L\setminus X$ with $X$? I'm not clear on what the more complex quantity is buying us?}

% \textcolor{red}{[TODO: Explain Figures 2b and 2c with notation.]}

% \textcolor{red}{[TODO: Consider: in instruction-tuned models, is this really a "do" $X$ (or "do" $L$)? i.e., is it $L \rightarrow X$?]}

% \textcolor{red}{[TODO: Actually, maybe we just stick with Figure 1 and do b, c variants. Also add $L \rightarrow A$. Add red $X$ over $X \rightarrow A$ or something, some indicator of the fact that when you intervene on $A$, you think that this arrow is getting cut but it actually isn't]}

% \vspace{-0.3em}
\section{Detecting and adjusting for selection bias}
% In this section, we introduce approaches for identifying and mitigating selection bias in LLM user simulation experiments.

Having argued that synthetic experiments are inherently prone to selection bias, we provide in this section an empirical strategy to both diagnose and mitigate this bias in practice.
The broad idea is to treat the synthetic experiment as an observational study, but one where we have the opportunity to query additional variables for diagnosis and adjustment at analysis time.

\subsection{Detecting confounding with negative control outcomes}
% \label{sec:negative-controls}

% Our analysis suggests that synthetic user studies can suffer from a failure of comparability: even when the explicit persona specification $L$ is held fixed, different intervention conditions $A$ can induce different distributions over unspecified latent attributes $X\setminus L$, so that
% \begin{equation}
%     P(X\setminus L \mid A=1, L) \neq P(X\setminus L \mid A=0, L).
%     \label{eq:latent-shift}
% \end{equation}
% When this occurs, the contrast in outcomes across intervention conditions reflects not only the causal effect of $A$ on $Y$ but also a shift in the latent user population being implicitly sampled.

% In this section 

% Our analysis suggests that synthetic user studies can experience bias due to user drift. 
Because the underlying persona $X$ of a synthetic user is not observed in full, user drift and selection bias cannot be directly computed. Instead, we develop a diagnostic for confounding using targeted \textit{negative control outcomes} $Z$ \citep{lipsitch2010negative}, which serve as observed proxies for the unmeasured $X \setminus L$. In our usage, negative control outcomes should depend on the same unspecified latent user attributes $X \setminus L$ whose drift induces confounding bias in the primary outcome, but should not be \emph{causally} affected by the intervention condition $A$ (Figure \ref{fig:dag:negative-control}). 
% \john{I thought negative controls typically depended on $L$ and $X$, i.e. the same dependencies as $A$?.}
Then as with $P(X \mid A=a, L)$, in the absence of confounding the negative control outcome distribution $P(Z \mid A=a, L)$ should remain invariant to $A$, i.e., it should hold that $P(Z \mid A=1, L)=P(Z \mid A=0, L)$.

At a high level, a good negative control outcome satisfies the following two properties:
\begin{enumerate}
    \item \textit{No causal effect of $A$ on $Z$}: In an actual randomized experiment, $Z$ should not change under interventions on $A$ (given $L$).
    \item \textit{Sensitivity to latent user attributes}: $Z$ should depend on (or be informative about) the unspecified latent user attributes $X\setminus L$ that may differ across intervention conditions.
    % \john{Isnt this typically $X$-comparable to $A$, i.e. depends on $L$ and $X$ ? Seems to be this way in figure 1b}
\end{enumerate}

Concretely, we consider outputs to prompts that target stable user traits not specified by $L$ (e.g., demographics, self-reported beliefs) as negative control outcomes.
% In the real-world data-generating process (Figure 1b), these variables are stable pre-treatment covariates that casually influence the primary outcome $Y$. However, because an LLM generates these attributes after exposure to the intervention prompt, they function as post-treatment outcomes in our simulation. By selecting covariates that should theoretically remain invariant to the intervention $A$, we repurpose them as negative control outcomes to detect shifts in $X \setminus L$. 
Since these attributes should theoretically be invariant to the intervention $A$ in a real-world experiment (as they would be causally upstream), we can measure empirical shifts in $Z$ in a synthetic experiment to capture the user drift over intervention conditions induced by the LLM. We 
% then measure the extent to which the distribution of $Z$ varies across $A$ within each persona specification $L$,
quantify this
using the total variation distance (TVD) between the negative control outcome distributions for each persona specification $L$:
{%
\setlength{\abovedisplayskip}{4pt}%
\setlength{\belowdisplayskip}{4pt}%
\setlength{\abovedisplayshortskip}{4pt}%
\setlength{\belowdisplayshortskip}{4pt}%
\begin{equation*}
    \text{TVD}=0.5\sum_Z \; \bigl|P(Z \mid A=1,L) - P(Z \mid A=0, L)\bigl|
\end{equation*}
}%
The magnitude of the TVD can be interpreted as evidence of how strongly the intervention conditions correspond to different latent user populations, providing an indication of the extent to which the observed effect is vulnerable to confounding bias. We note that because empirical TVD is positively biased in finite samples, it will naturally be bounded away from zero due to sampling variance alone.

% In principle, a non-zero TVD can be interpreted as evidence that the intervention conditions correspond to different latent user populations and therefore that the observed effect is vulnerable to confounding bias. We note, however, that because empirical TVD is positively biased in finite samples, it will be bounded away from zero due to sampling variance alone, and so we instead consider the relative magnitude of the TVD.
% Consequently, our adjustment procedure relies on a strictly positive stopping threshold $\epsilon$ that accounts for this statistical artifact.
    
% An important note is that t
The usefulness of TVD as a metric to quantify confounding is of course sensitive to one's choice of negative control outcomes. TVD scales with the amount of confounding only if the negative control outcome is chosen to be sufficiently sensitive to $X\setminus L$ and sufficiently insensitive to $A$. Furthermore, if we choose negative control outcomes that are only weakly related to $X\setminus L$, then we may appear to have less selection bias than we really do. In later sections, we discuss and provide evidence of this.

% \paragraph{Operationalizing negative controls in LLM user simulations.}
% In practice, we instantiate negative controls by eliciting responses that (i) reveal latent attributes or stable preferences, but (ii) should not change when only the intervention condition $A$ changes. Concretely, we consider prompts or scoring functions that target stable user traits (e.g., background preferences, self-reported beliefs, or style-invariant knowledge) and treat their outputs as negative control outcomes $Z$. We then measure the extent to which the distribution of $Z$ varies across $A$ within each persona specification $L$.

% \paragraph{Interpreting violations.}
% A detected shift in $Z$ can be interpreted as evidence that the two intervention conditions correspond to different latent populations, and hence that the primary outcome contrast is vulnerable to backdoor-induced bias. Importantly, this diagnostic does not require observing $X\setminus L$ directly; it only requires a negative control outcome that is sufficiently sensitive to $X\setminus L$ and sufficiently insensitive to $A$. The next section builds on this diagnostic by using additional elicited attributes $L'$ to reduce the induced population shift, aiming to make the comparison across $A$ closer to a randomized experiment within strata of $(L,L')$.

% \vspace{-0.3em}
\subsection{Adjusting for confounding}
\label{sec:adjustment}

\begin{algorithm}[t]
\caption{Iterative detection and confounder adjustment heuristic}
\label{alg:proxy-adjustment}
% \small
\begin{algorithmic}[1]
\Require Treatment $A\in\{0,1\}$; personas $\{L_i\}_{i=1}^n$; trials per persona $T$; TVD threshold $\epsilon > 0$
\Ensure Elicited confounders $\{L_i'\}_{i=1}^n$ and primary/negative control outcomes
\State Initialize augmented persona $\tilde L_i \gets L_i$ and $L_i' \gets \emptyset$ for each $i \in \{1 \dots n\}$
\Repeat
    \For{each persona $i \in \{1 \dots n\}$, condition $a \in \{0,1\}$, and trial $t \in \{1 \dots T\}$}
        \State Simulate interaction under $A\!=\!a$ with $\tilde L_i$; record $Y_{i,t,a}$ and $Z_{i,t,a}$
    \EndFor
    
    \State Compute $\mathrm{TVD}_i \gets \tfrac{1}{2}\sum_Z\;\bigl|P(Z\mid A=1,\tilde L_i)-P(Z\mid A=0,\tilde L_i)\bigr|$ for each $i$
    
    \If{$\frac{1}{n}\sum_{i=1}^n \mathrm{TVD}_i > \epsilon$}
        \For{each persona $i \in \{1 \dots n\}$, condition $a \in \{0,1\}$, and trial $t \in \{1 \dots T\}$}
            \State Elicit additional attributes $L'_{i,t,a} \sim P(L'\mid A=a,\tilde L_i)$
        \EndFor
        
        \For{each persona $i \in \{1 \dots n\}$}
            \State Sample one elicited attribute set $L_i' \gets \mathrm{Unif}\bigl(\{L'_{i,t,a}: t\in[T], a\in\{0,1\}\}\bigr)$
            \State Update augmented persona $\tilde L_i \gets (\tilde L_i \cup L_i')$
        \EndFor
    \EndIf
\Until{$\frac{1}{n}\sum_{i=1}^n \mathrm{TVD}_i \le \epsilon$ (or a max-iterations budget is reached)}
\end{algorithmic}
\end{algorithm}
% \vspace{-0.5em}

The negative control diagnostic can detect confounding in synthetic experiments, but it does not by itself specify how to reduce the resulting bias. Here, we propose a practical adjustment strategy: elicit additional attributes $L'$ that proxy the unspecified latent attributes $X\setminus L$, 
% \john{Aren't we choosing those attributes based on the observed changes in the persona specific to a particular $Z$? Could we use matching labels $Z_1$, $L_1$?} 
then instantiate a new simulated user for whom both the initial observed attributes $L$ \textit{and} the elicited attributes $L'$ are specified (Figure~\ref{fig:dag:adjust}).
% \footnote{Though this procedure approximates conditioning, adding an attribute to the context of an LLM is not necessarily the same as conditioning on it statistically. In particular, the type of conditioning that arises from an LLM \textit{holding} a prior belief may be different from the type of conditioning that arises from an LLM \textit{talking about holding} a prior belief.} Philosophically, we view this as akin to adjusting for confounders in a traditional observational causal inference setting.

% \john{I noticed that the current targeted confounders set ask for users opinions about various things, including immigration etc. I think that's problematic here. The LLMs latent state comprises a persona piece $X$ and a context (call it $C$) which includes the user's dynamic state. We're trying to fix $X$ to avoid confounding, which the demographic confounders should do nicely. But $C$ is supposed to be shaped by our intervention $A$, and we want to measure the full effect of that (i.e. $C$ is part of the causal mechanism we want to measure, not a confounder).  The leading questions should change the models answers, as they would for a human. We dont want to reduce that effect by pinning down the user's short-term opinion before $A$ is applied. i.e. we dont want to deconfound through $C$ and we dont want to see sudden drops in $Y$ at position 4 in figure 3, which are in evidence for some of the models. }

\textbf{Eliciting confounders.} \;
We draw new confounders for each persona from the treatment-marginal distribution of confounders, which is a mixture of the treatment-conditional confounder distributions:
% for each persona:
\begin{equation*}
    L' \sim 0.5 \cdot P(L' \mid A=1, L)  + 0.5 \cdot P(L' \mid A=0, L)
\end{equation*}
Note that this treatment-uniform mixture distribution is distinct from the observational distribution marginalized over treatment, $P(L' \mid L) = \sum_{a \in \{0,1\}} P(A=a \mid L)P(L'\mid A=a, L)$.
We draw from the treatment-uniform mixture to ensure that the additional confounders reflect the distribution that would be induced by running a real randomized experiment, where treatment conditions are assigned with equal probability.
% This---as in a real randomized experiment---prevents 
% By sampling from this mixture distribution, we aim to preserve the natural variance of the simulated population. 
The objective is to fix the elicited attributes $L'$ across intervention conditions
% , ensure that the attributes $L'$ that will be added to the persona specification are fixed across intervention conditions for a given user, thereby balancing the remaining latent attributes $X \setminus (L \cup L')$ across arms 
without narrowing the overall user distribution. 

% \textcolor{red}{[TODO: Explain why we sample $L'$ from $P(L'|A=a,L)$ instead of $P(L'|L)$. We're accounting for the fundamental likelihood of the treatment, i.e., we're trying to balance in the population that's been induced by the treatment. Explain relationship between the two distributions. "Note that this distribution is distinct from the marginal distribution $P(L'|L)=P(A|L)P(L'|A,L)$... in cases where one treatment condition is rare in the population, drawing from this mixture distribution draws from a more balanced sample w.r.t. the treatment assignment / matches the population induced by the experiment, which gives better external validity w.r.t. the experiment." "This is the distribution that would be induced by running the standard experiment"]}

In practice, we simulate $T$ trials for each user. In each trial, we prompt the user directly for a confounder value following intervention in each arm of the trial (e.g., by asking a targeted question). Across all trials for that user, we randomly select one of these realizations to add to the user persona for the next iteration, resulting in a single arm-independent $L'$ across all trials for a specific user.

\textbf{Causal caveats.} \; 
% \cmt{AD: Took a crack at this section (old text commented after each sentence. I think we could put this at the end of this section, and bring the final short paragraph under this header.}
Perfect causal adjustment aims to block backdoor paths by conditioning on variables that render treatment assignment independent of latent determinants of the outcome. Sampling and adjusting for $L'$ in the manner we propose does not guarantee this, in part because prompting an LLM with descriptions of attributes approximates but is not equivalent to statistically conditioning on them.
% Admittedly, we cannot guarantee that 
% Sampling $L'$ in the manner we propose does not guarantee this, and augmenting an LLM's prompt context with text descriptions of attributes approximates but is not equivalent to statistically conditioning on them. 
% will succeed at this.
%In our setting, $L'$ is not assumed to perfectly recover $X\setminus L$, and we do not claim full identification of $\tau^*$.
Rather, we aim for bias reduction:
we choose new confounders $L'$ such that the remaining unobserved persona attributes $X \setminus (L \cup L')$ 
% \john{Perhaps talk instead about the narrowed posterior $P(X | L, L')$ ?}
that influence outcomes are less imbalanced across treatment arms.
As with any observational study, we cannot guarantee success, but we can build falsifying evidence with well-chosen negative controls $Z$.
%, or provide quantitative caveats with sensitivity analysis [CITES].
%fixing $L'$ in addition to $L$ should reduce some of the variation in $X\setminus L$ (now $X \setminus (L \cup L')$) that differs across intervention conditions. 
% This helps to ensure that comparisons across $A$ are made over more similar latent user populations.

We draw attention to an important caveat: because we are unlikely to be able to fully specify $X\setminus L$ via $L'$ (i.e., there is always some unmeasured confounding), then 
% confounding bias does not necessarily decrease linearly in  
$L'$ can reduce bias but is not guaranteed to eliminate it, especially if substantial latent variation remains in $X\setminus (L \cup L')$ or if the elicitation procedure induces additional dependencies. This means that confounding bias also does not necessarily decrease monotonically in the size of $L \cup L'$.

\textbf{Iterative adjustment.} \;
In practice, rather than eliciting all possible confounders and providing them all at once in a detailed persona specification, we elicit small batches of attributes iteratively and add them to the persona specification, monitoring confounding at every iteration until the distance between the treatment-specific user populations is sufficiently reduced (Algorithm \ref{alg:proxy-adjustment}). 

The motivation for this iterative procedure is threefold. First, confounder elicitation is computationally expensive, and adding long lists of confounders to the conversational context as part of adjustment further increases inference overhead. Second, LLMs suffer from performance degradation given long context windows, and expanding the persona specification to include all possible confounders increases this risk. 
% \cmt{TY: could add saving context lengths and LLM's known performance degradation in long context queries as additional justification}. 
Third---because we ``condition'' not in the statistical sense but rather by sampling a fixed set of confounders that we specify in an LLM prompt---every constraint we add to the persona further runs the risk of narrowing the user population through sampling bias, which can affect the external validity of the effect computed from the synthetic experiment. Therefore, we attempt to identify a \textit{minimum} set of confounders that sufficiently controls for confounding bias in the outcome.

\section{Experiments}
\label{sec:experiments}
% \subsection{Experimental setup}
\textbf{Models.} \; 
We evaluate synthetic users simulated using the following diverse range of frontier models: Qwen3-30B-A3B (both fine-tuned and instruction-tuned versions) \citep{yang2025qwen3technicalreport}, GPT-OSS-20B \citep{openai2025gptoss120bgptoss20bmodel}, Gemma-4-31B (instruction-tuned) \citep{gemma4_2026}, and Gemini 3 Flash \citep{gemini3_api_2026}. We additionally include Gemma-3-4B (instruction-tuned) \citep{gemma3_4b_2026} to explicitly test the behavior of a smaller-capacity model. Model details are available in Appendix \ref{sec:appendix:model_details}.

% \begin{itemize}
%     \item Gemma-4-31B (instruction-tuned)
%     \item Qwen3-30B-A3B (both base and instruction-tuned versions)
%     \item GPT-OSS-20B
%     \item Gemma-3-4B (instruction-tuned)
%     \item Gemini 3 Flash
% \end{itemize}

\textbf{Procedure.} \; We begin by instantiating the synthetic user with base persona attributes $\tilde L=L$ derived from an existing dataset of users. In parallel conversations, identically instantiated users are presented with interventions $A=0$ and $A=1$. In each arm, the user is then asked---in diverging conversation branches---the primary outcome question (eliciting $Y$), a negative control question (eliciting $Z$), or a confounder question (eliciting $L'$). As each question is asked independently of the others, the simulated user sees \textit{only} that question and the intervention, not any of the other questions. These answers are then collected and used to compute both the observed effect and the TVD.\footnote{For questions where a list of options is provided, answers matching none of the options are mapped to \textit{Unknown}.} 
% \john{Isn't $Z$ normally attached to $X$ and $L$ and not $A$? If $Z$ is computed in a separate branch from $Y$, why is there an edge in your figure 1b graph from $Z$ to $Y$? Should you average the two $Z$ values in the $A$ branches using $P(A|L)$ (derivable from model likelihoods) to remove the arrow from $A$ to $Z$?}
Of the sampled confounders, a subset is added to the persona specification, and the procedure is repeated from the beginning with a more detailed persona instantiation $\tilde L=(\tilde L \cup L')$. The conversation history from the previous iteration is \textit{not} included in the new iteration.
% Our evaluations follow this general procedure:
% \begin{enumerate}
%     \item The synthetic user is instantiated with base persona attributes $L$ derived from an existing dataset of users.
%     \item The user is presented with an intervention $A$.
%     \item The user is asked, in parallel conversation branches:
%     \begin{itemize}
%         \item The primary outcome question.
%         \item A negative control question.
%         \item A confounder question.
%     \end{itemize}
%     \item These answers are collected and used to compute both the observed effect and the TVD. For questions where a list of options are provided, answers matching none of the options are mapped to \textit{Unknown}.
%     \item Of the sampled confounders, a subset are added to the persona specification. We then repeat the procedure from step 1.
% \end{enumerate}

% The primary outcome, negative control, and confounder questions all follow the intervention, but each question is asked independently of the others. That is, the simulated user does not see any of the other questions prior to being asked a specific question---only the intervention.

For each experimental setting, we sample 100 personas from the dataset to instantiate 100 corresponding synthetic users. We then conduct 30 independent trials per user, evaluating both intervention conditions in parallel during each trial. Due to computational constraints, Gemini 3 Flash experiments use 10 synthetic users instantiated from 10 sampled personas, with 5 trials per user.

\vspace{-0.05em}
\subsection{Simulated user survey}

We evaluate two types of settings where we realistically expect synthetic users to be used. The first is in survey-type experiments (e.g., in social science), where simulated users are asked to express their opinions or preferences in response to fixed questions.

For this setting type, we draw from the OpinionQA dataset \citep{pmlr-v202-santurkar23a}, which contains questions and human responses from the Pew American Trends Panel opinion polling surveys, 
% and individual human responses to each of those questions,
along with demographic attributes of the respondents. In this setting, simulated users are given a leading statement before being asked to respond to the survey questions in the OpinionQA dataset. We use the following experimental setup:
\begin{itemize}
    \item $L$: The age and sex of the respondent, seeded from the actual OpinionQA respondents.
    \item $A$: A leading statement to precede the survey question. Each intervention condition targets one of the possible answers to the survey question.
    \item $Y$: The respondent's answer to the survey question. We choose the question specifically to be one that suggests the political affiliation of the respondent.
    \item $Z$: The synthetic user's elicited citizenship status, political ideology, political party, race, and answers to 5 other survey questions that similarly provide an indication of political affiliation.
    \item $L'$: The synthetic user's elicited geographic region, income, level of education, marital status, religion, religious attendance frequency, and 20 targeted LLM-generated questions that specifically relate to the intervention and outcome (10 groups of 2 questions each). Confounders are added over iterations in groups of 2, in the order they are listed here.
\end{itemize}

% We sample 100 personas from the original dataset and conduct 30 trials for each. Due to computational constraints, Gemini 3 Flash experiments use 10 personas with 5 trials each. 
Prompts and lists of individual questions can be found in Appendix \ref{sec:appendix:shared_implementation_details}, \ref{sec:appendix:opinionqa_details}.

\vspace{-0.1em}
\subsection{Assistive agent evaluation}

A second type of setting where we expect simulated users to be used is in evaluating LLM agents, e.g., an A/B test of whether a new version of an LLM agent is better than an existing version. Here, simulated users might engage in a conversation with the agent before being prompted with a primary question relating to their interaction with the agent.

As a common example of this type of interaction, we take a setting where the simulated user engages with the LLM agent to ask for a recommendation. We examine two variations of this setting where the user asks for either a book or a movie. The agent responds, and the two converse autonomously for several turns before the user is then asked how likely they are to read or watch the book or movie being discussed. We use the following experimental setup:
\begin{itemize}
    \item $L$: The age and sex of the respondent, seeded from the actual respondents in the NYT Book Opinions\footnote{We use only the titles and authors of 5 books from this dataset and the annotators' demographic information, and not any content created by the New York Times.} \citep{meister-etal-2025-benchmarking} and MovieLens \citep{harper-2015-movielens} datasets.
    \item $A$: Different system prompts for the LLM agent (Gemma-4-31B-it). The agent is instructed to discuss a specific book/movie in a positively ($A=1$) or negatively ($A=0$) biased way.
    \item $Y$: How likely the respondent is to read the book/movie being discussed, on a scale from \textit{1 - Very unlikely} to \textit{4 - Very likely}.
    \item $Z$: The synthetic user's elicited citizenship status, political ideology, political party, and race. The respondent is also asked, ``Have you heard of the <book/movie> we are discussing before?''
    \item $L'$: The synthetic user's elicited geographic region, income, level of education, marital status, religion, religious attendance frequency, and 20 targeted LLM-generated questions that specifically relate to the intervention and outcome (10 groups of 2 questions each). Confounders are added over iterations in groups of 2, in the order they are listed here.
\end{itemize}

% Because of the increased inference overhead from multiple dialogue turns between the synthetic user and agent, we sample 30 personas from the original dataset and conduct 10 trials for each. 
Due to computational constraints, Gemini 3 Flash experiments 
% use 10 personas with 5 trials each and 
run for 6 iterations. Prompts and lists of individual questions can be found in Appendix \ref{sec:appendix:shared_implementation_details}, \ref{sec:appendix:multiturn_agent_details}.

% \vspace{-0.2em}
\section{Results and Discussion}
% Results figures

\begin{figure*}[!t]
  \centering
  \includegraphics[width=\textwidth]{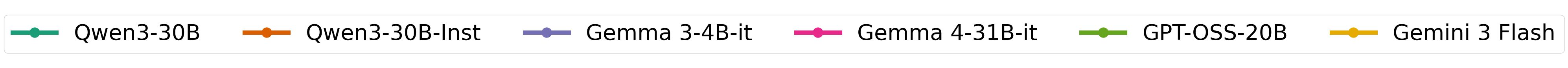}
  \begin{subfigure}[t]{0.33\textwidth}
    \centering
    \includegraphics[width=\linewidth]{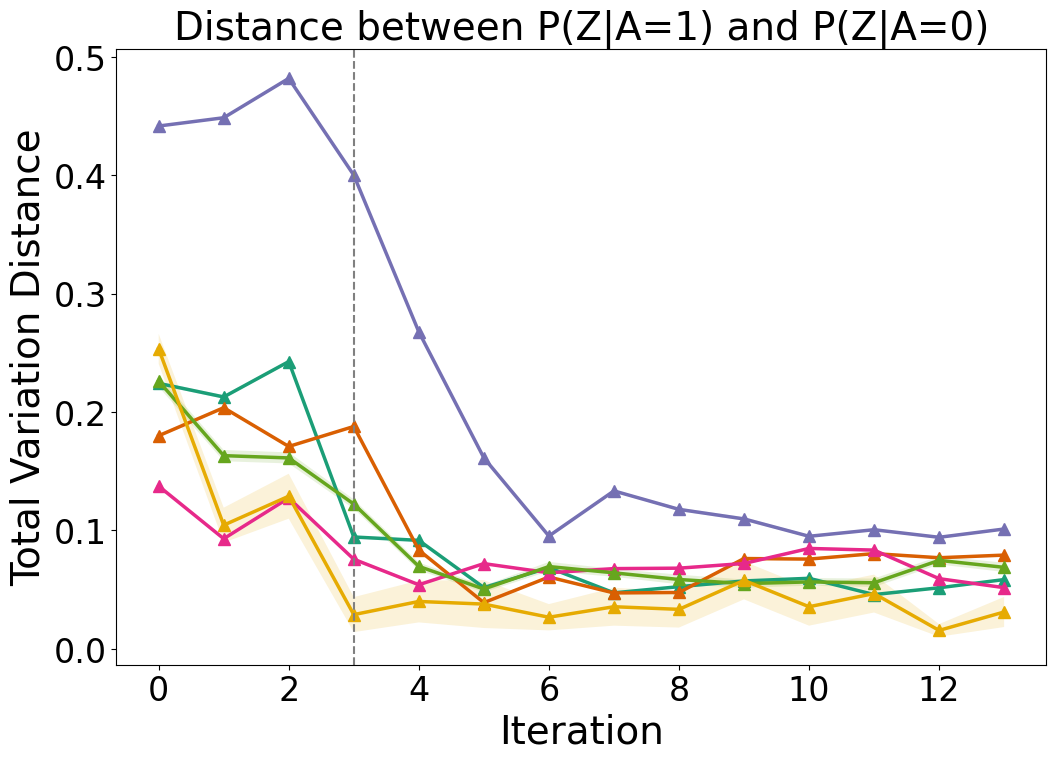}
    \caption{OpinionQA}
    \label{fig:tvd_opinionqa}
  \end{subfigure}\hfill
  \begin{subfigure}[t]{0.33\textwidth}
    \centering
    \includegraphics[width=\linewidth]{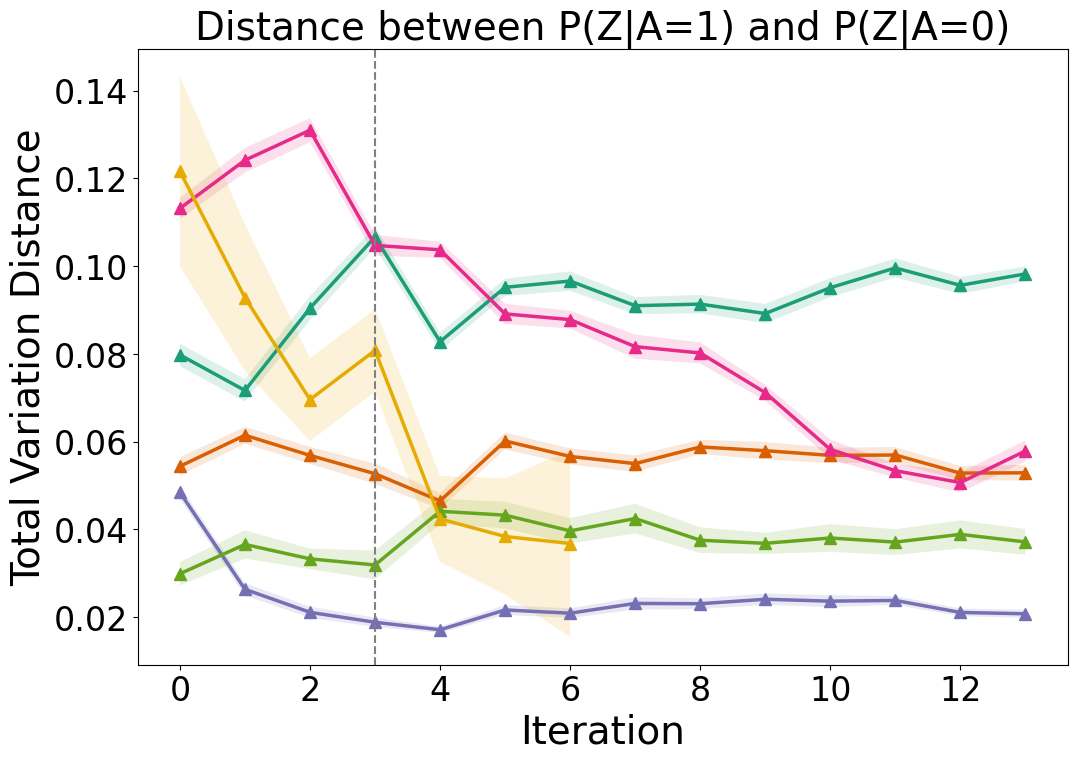}
    \caption{Book Opinions}
    \label{fig:tvd_nyt}
  \end{subfigure}\hfill
  \begin{subfigure}[t]{0.33\textwidth}
    \centering
    \includegraphics[width=\linewidth]{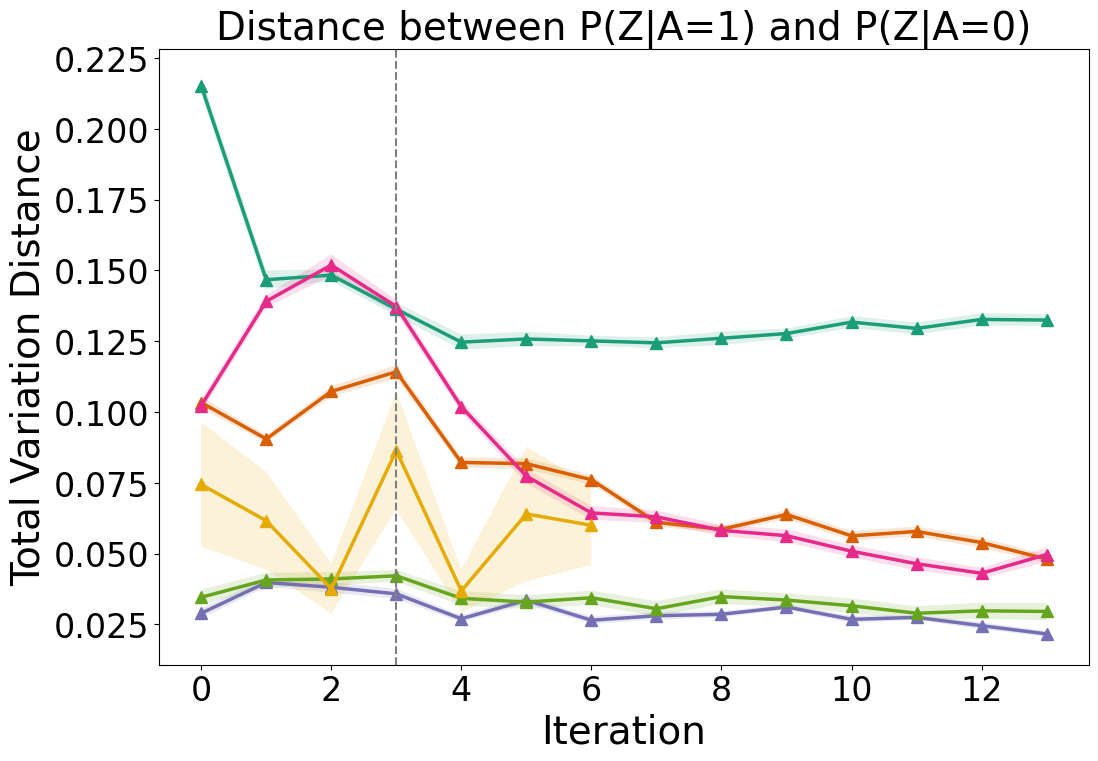}
    \caption{MovieLens}
    \label{fig:tvd_movielens}
  \end{subfigure}
  \caption{TVD over adjustment iterations. Shaded bands indicate 95\% CIs. 
  % Dashed line indicates the first step where targeted confounders are included in the persona specification in addition to demographic confounders.
  Dashed line indicates the switch from demographic to targeted confounders, i.e., the last step where only demographic confounders are included in adjustment.
  }
  \vspace{-1em}
  \label{fig:combined-tvd}
\end{figure*}
\begin{figure*}[!t]
  \centering
  % Trim order: left bottom right top
  \begin{subfigure}[t]{0.33\textwidth}
    \centering
    \includegraphics[width=\linewidth]{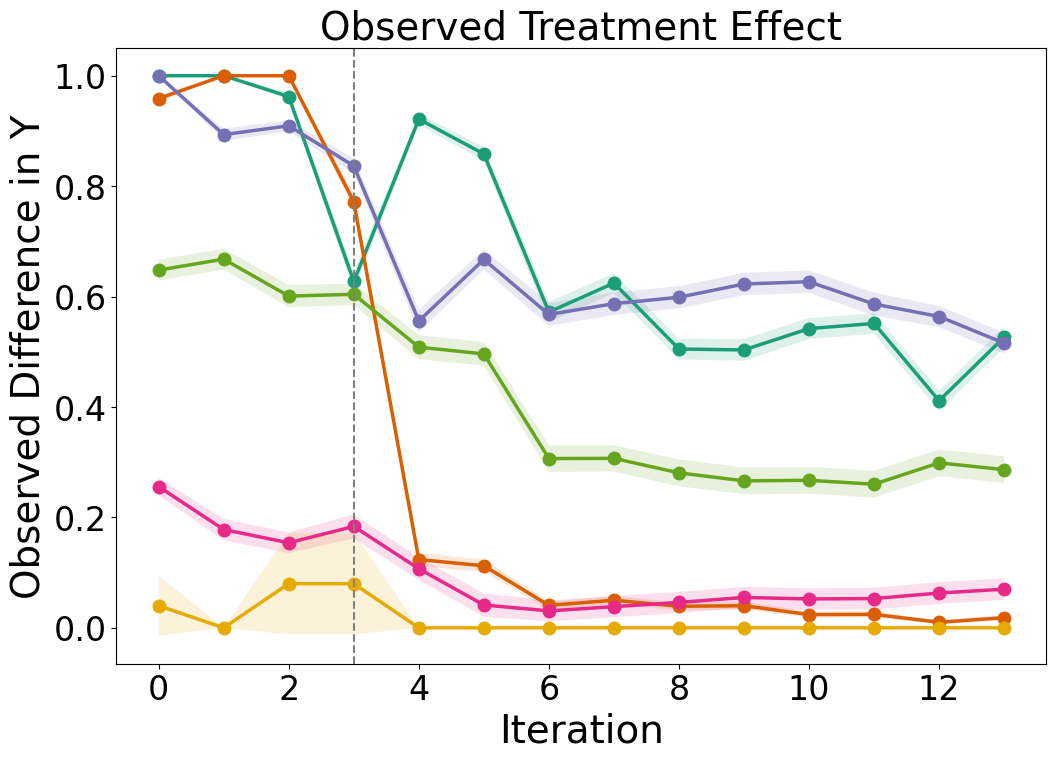}
    \caption{OpinionQA}
    \label{fig:effect_opinionqa}
  \end{subfigure}\hfill
  \begin{subfigure}[t]{0.33\textwidth}
    \centering
    \includegraphics[width=\linewidth]{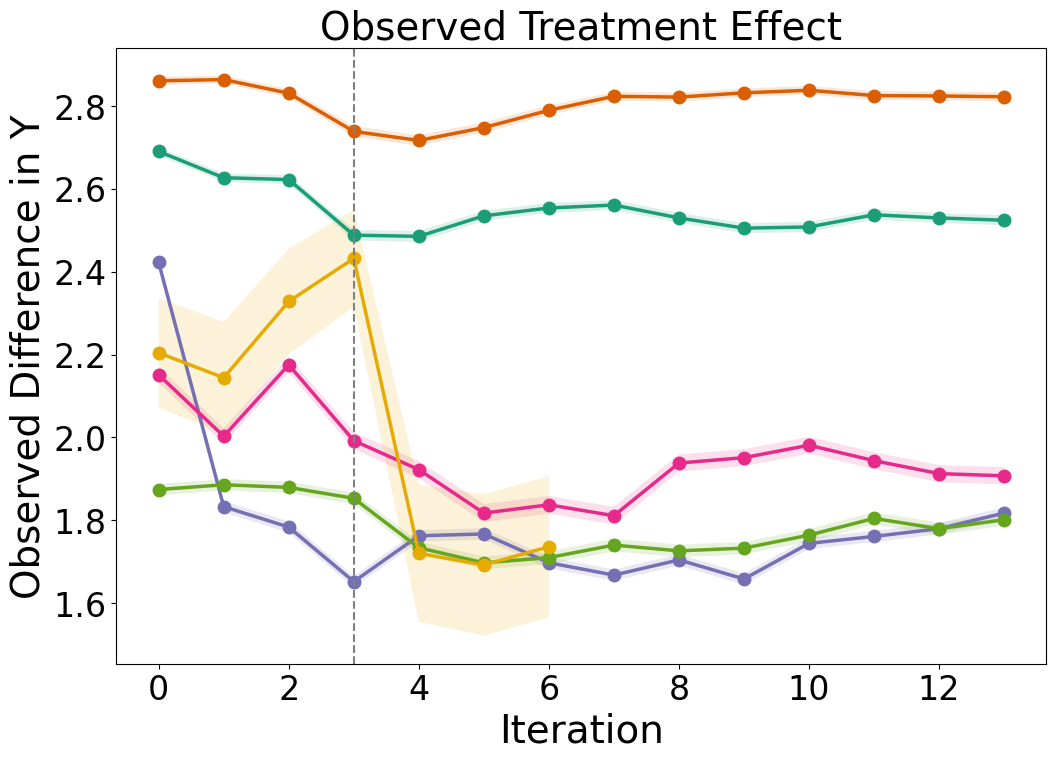}
    \caption{Book Opinions}
    \label{fig:effect_nyt}
  \end{subfigure}\hfill
  \begin{subfigure}[t]{0.33\textwidth}
    \centering
    \includegraphics[width=\linewidth]{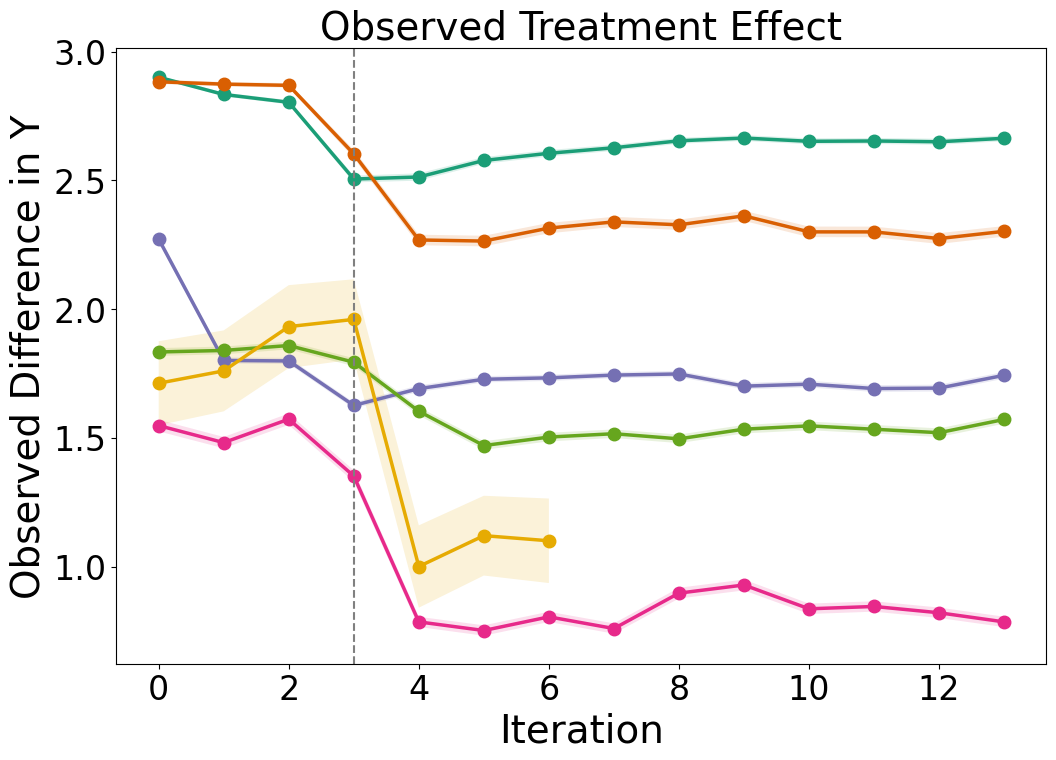}
    \caption{MovieLens}
    \label{fig:effect_movielens}
  \end{subfigure}
  \caption{Observed effects over adjustment iterations. Shaded bands indicate 95\% CIs. 
  % Dashed line indicates the first step where targeted confounders are included in the persona specification in addition to demographic confounders.
    Dashed line indicates the switch from demographic to targeted confounders, i.e., the last step where only demographic confounders are included in adjustment.
    }
  % \vspace{-0.5em}
  \label{fig:combined-observed-effect}
\end{figure*}

We report results and provide a unified set of takeaways. Over adjustment iterations, we add elicited confounders to the persona specification, and we record (i) TVD between negative control outcome distributions under the two intervention conditions (Figure \ref{fig:combined-tvd}) and (ii) the observed treatment effect (Figure \ref{fig:combined-observed-effect}). Additional analysis and results are available in Appendix \ref{sec:appendix:additional_results}. A comprehensive set of plots for all negative control outcomes at every adjustment step is available in Appendix \ref{sec:appendix:additional_plots}.

\vspace{-0.1em}
\subsection{User drift and selection bias are present}
\vspace{-0.1em}

% \textbf{Takeaway 1: Without adjustment, selection bias/user drift exists between intervention conditions.} \; 
% At 
% \textit{Without adjustment, selection bias/user drift exists between intervention conditions.} 
Before adjustment (iteration 0), we observe substantial TVD between negative control outcome distributions across models and datasets (Figure \ref{fig:combined-tvd}), indicating that the simulated user population under $A=0$ differs from that under $A=1$. In the survey setting, most models exhibit a similar amount of user drift between interventions with the exception of the small Gemma-3-4B, which has a particularly large TVD. In the agent evaluation setting, the same pattern holds for both Book Opinions and MovieLens, but with substantial variation in the magnitude of difference across models. 

We reiterate that \textit{selection bias is a natural consequence of a model with strong abductive reasoning abilities}. 
% Because models are trained to infer about unspecified elements of their context, differences in populations across interventions are a sign that the model has strong abductive reasoning capabilities and is functioning as intended. 
Notably, in the MovieLens setting, the two smallest and oldest models---GPT-OSS-20B and Gemma-3-4B, which we expect to have the weakest abductive capabilities---show very little difference in negative control outcomes across interventions.
% , which may be a function of their weaker abductive capabilities.

\subsection{Iterative confounder adjustment is (often) effective}

We observe that \textbf{iterative confounder adjustment appears to be effective in controlling user drift and selection bias for many models, but not all.}
% \textbf{Takeaway 2: Iterative confounder adjustment appears to be effective in controlling selection bias/user drift for many models, \textit{but not all}.} \;
In the survey setting, TVD decreases consistently over iterations, 
% as additional elicited attributes $L'$ are incorporated in the persona specification,
suggesting that the adjustment procedure can reduce user drift (Figure \ref{fig:tvd_opinionqa}). In the agent evaluation setting, many model-dataset pairs exhibit clear reductions in TVD (e.g., Gemma-4-31B in both Book Opinions and MovieLens, Gemini 3 Flash and Gemma-3-4B in Book Opinions, and both fine-tuned and instruction-tuned Qwen3 in MovieLens), but others (primarily GPT-OSS-20B) show flat or weakly changing TVD across iterations (Figures \ref{fig:tvd_nyt}, \ref{fig:tvd_movielens}). These results suggest that prompt-based conditioning and the particular elicited attributes may not be uniformly effective for confounding adjustment in multi-turn settings.

% \textbf{Takeaway 3: User drift affects the observed effect estimates.} \;
Additionally, \textbf{user drift affects the observed effect estimates.}
In both settings, the observed treatment effect changes over early adjustment iterations and then stabilizes after several iterations (Figure \ref{fig:combined-observed-effect}).
Because conditioning on additional attributes should not substantially change the observed effect if the user distributions between interventions are already comparable, these dynamics empirically support that (i) intervention-dependent user drift contributes to confounding bias in naive estimates and (ii) adjustment reduces the relevant sources of drift. Stabilization does not guarantee that all bias has been eliminated, but it suggests that additional confounder elicitation beyond that point (at least of related attributes) is less likely to address drift that materially changes the observed effect. 

% Additionally, the reduction in the observed effect over adjustment iterations highlights one of the dangers of user drift: misleading observed effects that over-attribute improvements to the intervention.
% suggests that 
% confounding bias was inflating the apparent treatment effect. W
% without adjustment, this can yield misleading conclusions that over-attribute improvements or advantages to one intervention condition relative to the true causal effect.
% The observed effect for Gemini 3 Flash in the survey setting behaves anomalously, beginning close to 0 and converging to 0 after several iterations (Figure \ref{fig:effect_opinionqa}).
Gemini 3 Flash exhibits notably anomalous behavior for its OpinionQA observed effect, which begins close to 0 and after several iterations converges to 0 (Figure \ref{fig:effect_opinionqa}). 
While
this may be partially due
% we may attribute this partially 
to the smaller sample size for this model (10 personas with 5 trials each), it is more likely that this behavior arises from the additional training proprietary models like Gemini undergo to express ``neutral points-of-view'' on potentially sensitive topics \citep{geminiteam2025geminifamilyhighlycapable}. 
% \textcolor{red}{[POST-SUBMISSION: This raises an interesting tension because the TVD is still decreasing with adjustment, which suggests that the user populations are different even if their responses are not. So there's a question of what kinds of weird user distributions were induced during training that managed to be invariant to interventions on these particular topics.]}

% \begin{figure}[!t]
%   \centering
%   % \begin{subfigure}[t]{0.4\linewidth}
%   \begin{subfigure}[t]{0.45\linewidth}
%     \vspace{0pt}
%     \centering
%     % \begin{minipage}[t][2cm][t]{\linewidth}
%       % \centering
%     \includegraphics[width=\linewidth,trim=0cm 51.5cm 59cm 7cm,clip]{figures/nyt-100p-30t/Qwen-Qwen3-30B-A3B_nc_demo.png}
%     % \end{minipage}
%     \caption{Demographic}
%     \label{fig:ncdemo_nyt_qwen3}
%   \end{subfigure}
%   % \begin{subfigure}[t]{0.22\linewidth}   
%   \begin{subfigure}[t]{0.27\linewidth}
%     \vspace{0pt}
%     \centering
%     % \begin{minipage}[t][4.54cm][t]{\linewidth}
%     % \begin{minipage}[t][3cm][t]{\linewidth}
%     % \centering
%     \includegraphics[width=\linewidth,trim=0cm 54.5cm 36cm 5.5cm,clip]{figures/nyt-100p-30t/Qwen-Qwen3-30B-A3B_nc_curated.png}
%     % \end{minipage}
%     \caption{Targeted}
%     \label{fig:nccurated_nyt_qwen3}
%   \end{subfigure}
%   \vspace{-0.1em}
%   \caption{Negative control outcome distributions for Qwen3-30B on Book Opinions.}
%   \vspace{-0.8em}
% \end{figure}

\begin{figure}[!t]
  \centering
  % Trim order: left bottom right top
  \includegraphics[width=\textwidth,trim=0mm 0mm 0mm 0mm,clip]{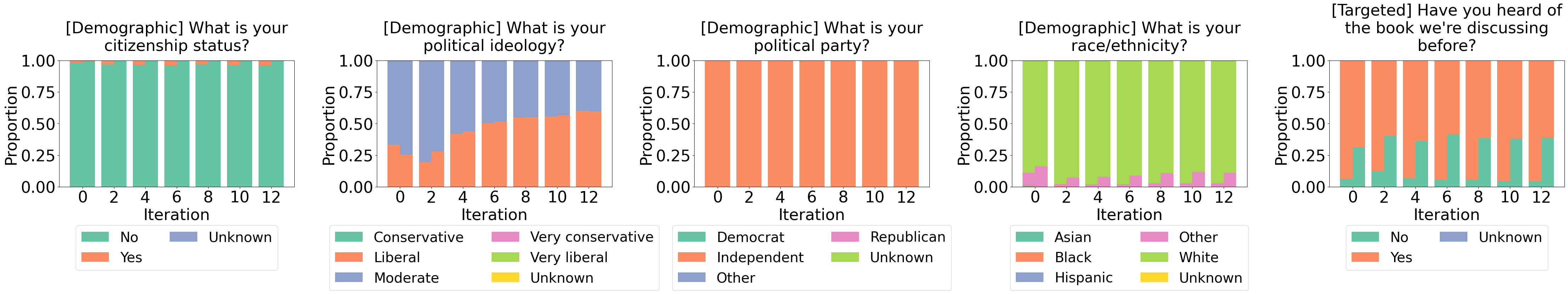}
  \caption{Negative control outcome distributions (Qwen3-30B, Book Opinions). At each adjustment iteration, $A=0$ is on the left and $A=1$ is on the right.}
  \label{fig:nc_nyt_qwen3}
  % \vspace{-1em}
\end{figure}

\vspace{-0.15em}
\subsection{Design choices are important}

% \textbf{Takeaway 4: Elicited confounders must be chosen strategically.} \;
\textbf{Confounders.} \;
In statistical causal inference, partial adjustment for certain covariates has the potential to worsen rather than remedy confounding. Akin to this phenomenon, several models (e.g., Gemma-3, Qwen3) in the survey setting exhibit an initial \textit{increase} in TVD before an eventual reduction. This increase corresponds to the iterations eliciting and adjusting for several generic demographic-based attributes: geographic region, income, education, and marital status. Subsequently, the simulated user's religious beliefs and habits are elicited and included in adjustment---which in contrast to the other demographic attributes does seem to decrease TVD across all three settings in many models. Finally, the elicited attributes then become the simulated user's answers to targeted questions related specifically to the intervention and outcome (this transition point is indicated by the vertical dashed lines in Figure \ref{fig:combined-tvd}). After these confounders are introduced in the persona specification, TVD continues to drop before stabilizing in later iterations.

This observation highlights a practical trade-off when constructing synthetic user populations. In persona instantiation, researchers typically rely on generic demographic attributes to span a target population with specific attributes and capture diverse user perspectives. The initial iterations of our adjustment loop show the impact of this practice: 
% To evaluate the impact of this practice, we introduce several of these generic attributes during the initial iterations of our adjustment loop. 
% As our results show, 
specification of generic attributes can in fact increase selection bias between intervention conditions. Since specifying such attributes is often necessary to define a sufficiently broad simulated population, our results suggest that a good practice is then to add targeted, task-specific confounders to subsequently bring selection bias back down.
% while adding generic attributes helps define a broad simulated population, it can inadvertently exacerbate selection bias between intervention conditions. Conversely, adding targeted, task-specific confounders reduces that bias. 
% Therefore, while including generic user attributes are often necessary for population representation, estimating unbiased effects instead requires more targeted adjustment.

% This observation suggests that in this partial adjustment setting, choosing confounders strategically is important to avoid increasing confounding bias. Given these constraints, conditioning on a small set of targeted, task-specific confounders appears to be more effective for reducing selection bias than conditioning on more generic attributes.

% \textbf{Takeaway 5: Negative control outcomes must be chosen strategically.} \;
\textbf{Negative control outcomes.} \;
Our selection bias diagnostic depends on the sensitivity of our chosen negative control outcomes to drift in the latent user attributes. In the Book Opinions dataset, for instance, negative control outcomes based on demographic attributes can show limited differences across intervention conditions for some models (Figure \ref{fig:nc_nyt_qwen3}) and therefore are not particularly effective proxies for shifts in $X \setminus L$. This can yield low or flat TVD even when the observed effect continues to change across iterations.
More targeted negative control outcomes, on the other hand, may be sensitive to the intervention, as seen for the rightmost negative control outcome in Figure \ref{fig:nc_nyt_qwen3}. Additionally, certain models like GPT-OSS exhibit high rates of refusal to answer in some data settings (Figure \ref{fig:ncdemomarginal_movielens_gptoss}, where refusal maps to \textit{Unknown}). If refusal dominates the negative control outcome distribution, then this may impact the TVD metric's ability to detect user drift.

% Moreover, certain models like GPT-OSS exhibit high rates of refusal to answer many of the negative control questions. Looking for instance at the MovieLens marginal negative control outcomes of GPT-OSS and Gemma-4 in Figures \ref{fig:ncdemomarginal_movielens_gptoss} and \ref{fig:ncdemomarginal_movielens_gemma4}, respectively, we see that the "Unknown" answer---which is our mapping of any answer that is not one of the provided options, but in our analysis typically corresponds to a model's refusal to provide an answer---is much more common for GPT-OSS. If refusal dominates the negative control outcome distribution, then this may impact the TVD metric's ability to detect user drift.

% \input{sections/additional_analysis}

% \vspace{-0.25em}
\section{Conclusion}
LLM-simulated user studies can undermine the internal validity we expect from randomized experiments, inducing confounding bias. In particular, interventions can induce shifts in latent user attributes such that outcomes for each intervention are effectively drawn from different user populations, leading to confounded effect estimates. We formalize this phenomenon as selection bias in synthetic experiments, propose negative control outcomes as a practical diagnostic, and empirically find evidence of such shifts across survey-style and multi-turn agent evaluations. Finally, we show that eliciting and specifying targeted user attributes during persona instantiation can substantially reduce observed user drift and stabilize effect estimates. Our findings suggest that synthetic user experiments should be treated as observational by default and that diagnosing and mitigating user drift in such experiments is essential for drawing robust conclusions about the causal effect.

% Open questions/future work
% \begin{itemize}
%     \item "Rolling forward" counterfactuals. In some sense we are already doing this since we set $L=(L,L')$ per person while selecting from a random treatment condition for that person, and we simulate both treatment conditions, so every person is going to have one arm where they have $L'|A=a$ traits while simulating $L'|A=1-a$. So why not do it fully?
%     \item Is it possible that we could "personalize" adjustment following notions of e.g. consistency, where for each person we sample confounders until their own personal $P(Z|A=1,L)=P(Z|A=0,L)$? This might mean that each person's $L'$ are different, at least in quantity (or even in what $L'$ are sampled, but that might be a harder sell...)
% \end{itemize}

\begin{ack}
% \textcolor{red}{[TODO]}
% Use unnumbered first level headings for the acknowledgments. All acknowledgments

% Do {\bf not} include this section in the anonymized submission, only in the final paper. You can use the \texttt{ack} environment provided in the style file to automatically hide this section in the anonymized submission.
This work was funded by Google LLC and/or a subsidiary thereof (Google DeepMind). VL, TY, MM, JC, AG, and AD are employees of Google LLC and may own Alphabet stock as part of a standard compensation package. All computational resources were provided by Google LLC.
\end{ack}

% \section*{References}

\clearpage
\printbibliography

%%%%%%%%%%%%%%%%%%%%%%%%%%%%%%%%%%%%%%%%%%%%%%%%%%%%%%%%%%%%

\clearpage
\appendix

% \textcolor{red}{[TODO: Don't forget to uncomment the appendix with all the treatment-conditional negative control plots]}
\section{Decomposition of the treatment effect}
% \textcolor{red}{[TODO: Could potentially be combined with section 2 and both made more compact]}
\label{sec:ate_decomposition}

In this section, we show theoretically how intervention-dependent user drifts can bias the observed effect $\tau$ between interventions. 
% We relate $\tau$ to the causal estimand $\tau^{\text{ATE}}$ and define terms that isolate how interventions can shift latent user attributes. This yields a decomposition of the observed difference as a sum of the true causal effect and a selection bias term that is specifically induced by user drift.
% relate the quantity measured in a naive synthetic experiment to standard causal estimands.
Given a persona specification $L$, and again working within the potential outcomes framework, let
\begin{equation*}
    \mu_{a,a'} = \int_{\mathcal{X}} \mathbb{E}[Y(a) \mid X, L]P(X \mid A=a', L)\; dX
\end{equation*}

Using this notation, we can rewrite several key quantities. In a synthetic experiment, the observed effect $\tau^{\text{obs}}$ for an LLM-simulated user with attributes $L$ is given by:
\begin{equation*}
    \tau^{\text{obs}}=\mu_{1,1}-\mu_{0,0}
\end{equation*}

The causal estimand of interest, the CATE, admits the standard decomposition into the average treatment effect on the treated (ATT) and on the controls (ATC).
% \begin{equation*}
% \begin{split}
%     \tau^{\text{ATE}}
%     & = P(A=1\mid L)\,\underbrace{\mathbb{E}[Y(1)-Y(0)\mid A=1,L]}_{\text{ATT}} 
%      \\
%     & \quad + P(A=0\mid L)\,\underbrace{\mathbb{E}[Y(1)-Y(0)\mid A=0,L]}_{\text{ATC}}
% \end{split}
%     \label{eq:ate-att-atc}
% \end{equation*}
When $P(A=1\mid L)=P(A=0\mid L)=1/2$, this becomes
% \begin{equation*}
%     2\tau^{\text{ATE}}(L)
%     = \bigl(\mu_{1,1}(L)-\mu_{0,1}(L)\bigr) + \bigl(\mu_{1,0}(L)-\mu_{0,0}(L)\bigr),
%     \label{eq:2tau-att-atc}
% \end{equation*}
\begin{align*}
    2\tau^{\text{ATE}}
        &=\underbrace{\mu_{1,1}-\mu_{0,1}}_{\text{ATT}}+\underbrace{\mu_{1,0}-\mu_{0,0}}_{\text{ATC}}
\end{align*}
We can also define the \textit{selection bias in the treated} (SBT) and the \textit{selection bias in the controls} (SBC). Intuitively, this is the difference in $Y$ resulting solely from user drift over $P(X \mid A=a,L)$ due to $A$.
\begin{equation*}
\delta_{\text{SBT}} = \mu_{1,1}-\mu_{1,0} \quad \quad 
\delta_{\text{SBC}} = \mu_{0,1}-\mu_{0,0}
\end{equation*}
Then we find that the CATE can be written in terms of the observed effect and the selection bias terms:
\begin{equation*}
    \begin{split}
        2\tau^{\text{ATE}}
            % &= \mu_{1,1}+\mu_{1,0}-\mu_{0,1}-\mu_{0,0} \\
            &=2\cdot(\underbrace{\mu_{1,1}-\mu_{0,0}}_{\tau^{\text{obs}}})-(\underbrace{\mu_{0,1}-{\mu_{0,0}}}_{\delta_{\text{SBC}}}) - (\underbrace{\mu_{1,1}-\mu_{1,0}}_{\delta_{\text{SBT}}})
    \end{split}
\end{equation*}
Letting average selection bias $\delta_{\text{SB}}=(\delta_{\text{SBC}}+\delta_{\text{SBT}})/2$,
\begin{equation*}
    \tau^{\text{ATE}}=\tau^{\text{obs}}-\delta_{\text{SB}}
\end{equation*}

% \cmt{TY: if we're tight on space, we can potentially keep the above equation only and move the derivation before that to Appendix}
This decomposition explicitly demonstrates that in experiments with LLM-simulated users, the observed effect will not correspond to the true causal treatment effect.
% , and the selection bias terms isolate which parts of the observed outcome difference in synthetic experiments can be attributed to user drift.
Depending on the direction of the selection bias, this can either artificially inflate or reduce the observed effect relative to the true causal effect, yielding misleading results.

\section{Additional results}
\label{sec:appendix:additional_results}
% \subsection{Additional discussion and analysis}

% \textcolor{red}{[VL: This section will probably be moved to the appendix]}

\begin{figure*}[!h]
  \centering
  % Crop each plot to the top-left facet (step 0). Adjust trim values if the plot layout changes.
  \begin{subfigure}[t]{0.32\textwidth}
    \centering
    \includegraphics[width=\linewidth,trim=0cm 28.7cm 36.7cm 0cm,clip]{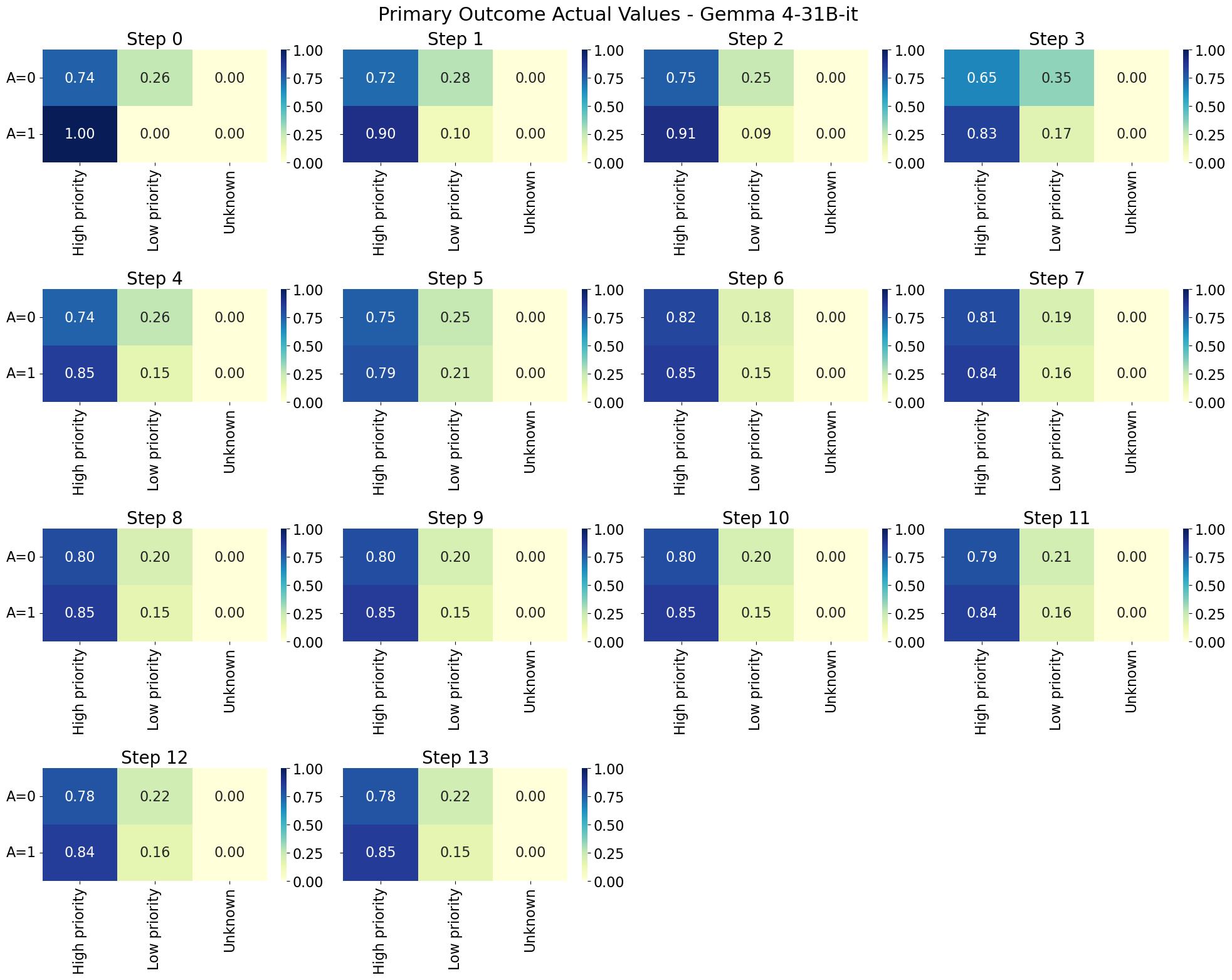}
    \caption{OpinionQA (Gemma-4-31B-it)}
    \label{fig:outcome_opinionqa_gemma4-it}
  \end{subfigure}\hfill
  \begin{subfigure}[t]{0.32\textwidth}
    \centering
    \includegraphics[width=\linewidth,trim=0cm 28.7cm 36.7cm 0cm,clip]{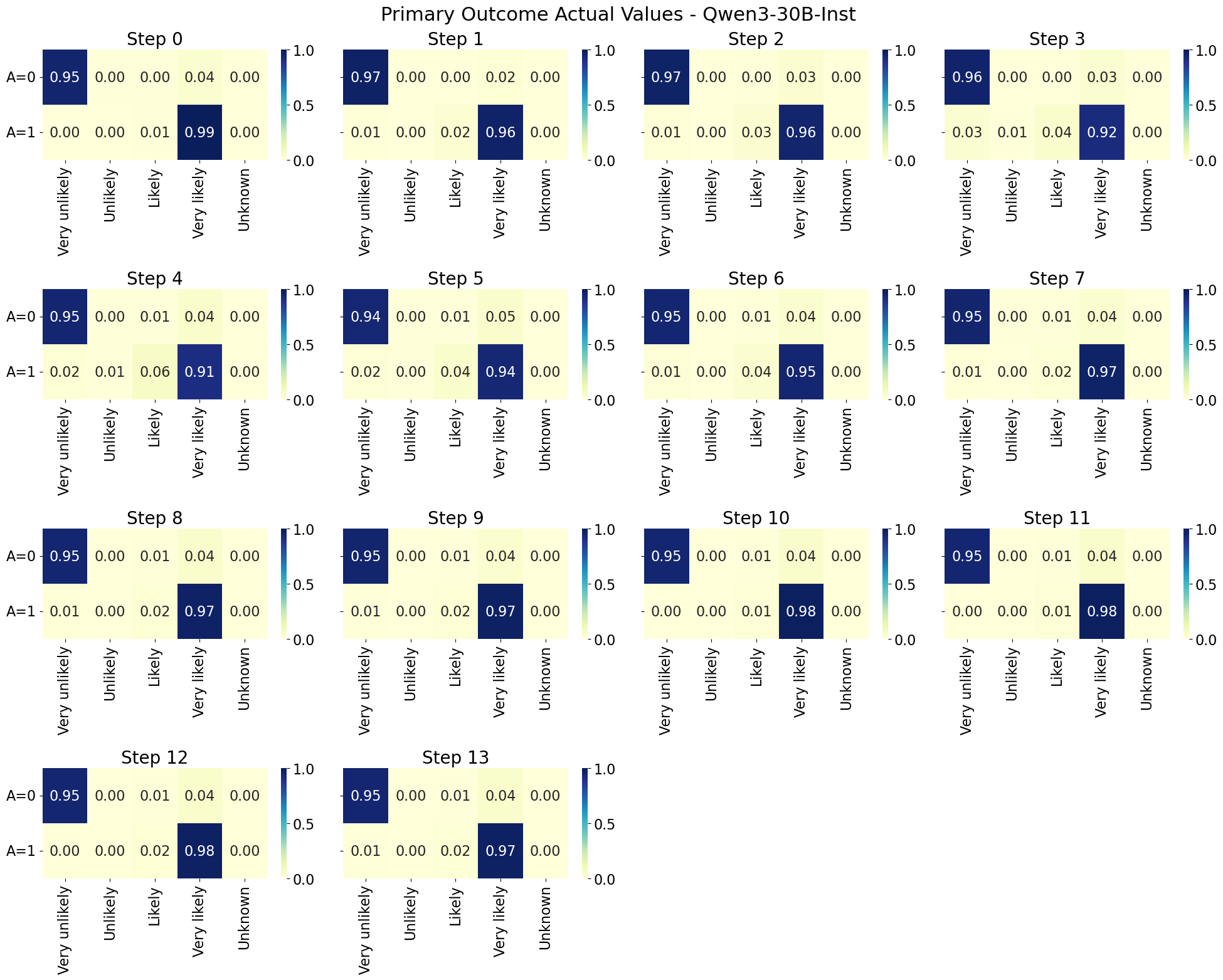}
    \caption{Book Opinions (Qwen3-30B-Instruct)}
    \label{fig:outcome_nyt_qwen3-instruct}
  \end{subfigure}\hfill
  \begin{subfigure}[t]{0.32\textwidth}
    \centering
    \includegraphics[width=\linewidth,trim=0cm 28.7cm 36.7cm 0cm,clip]{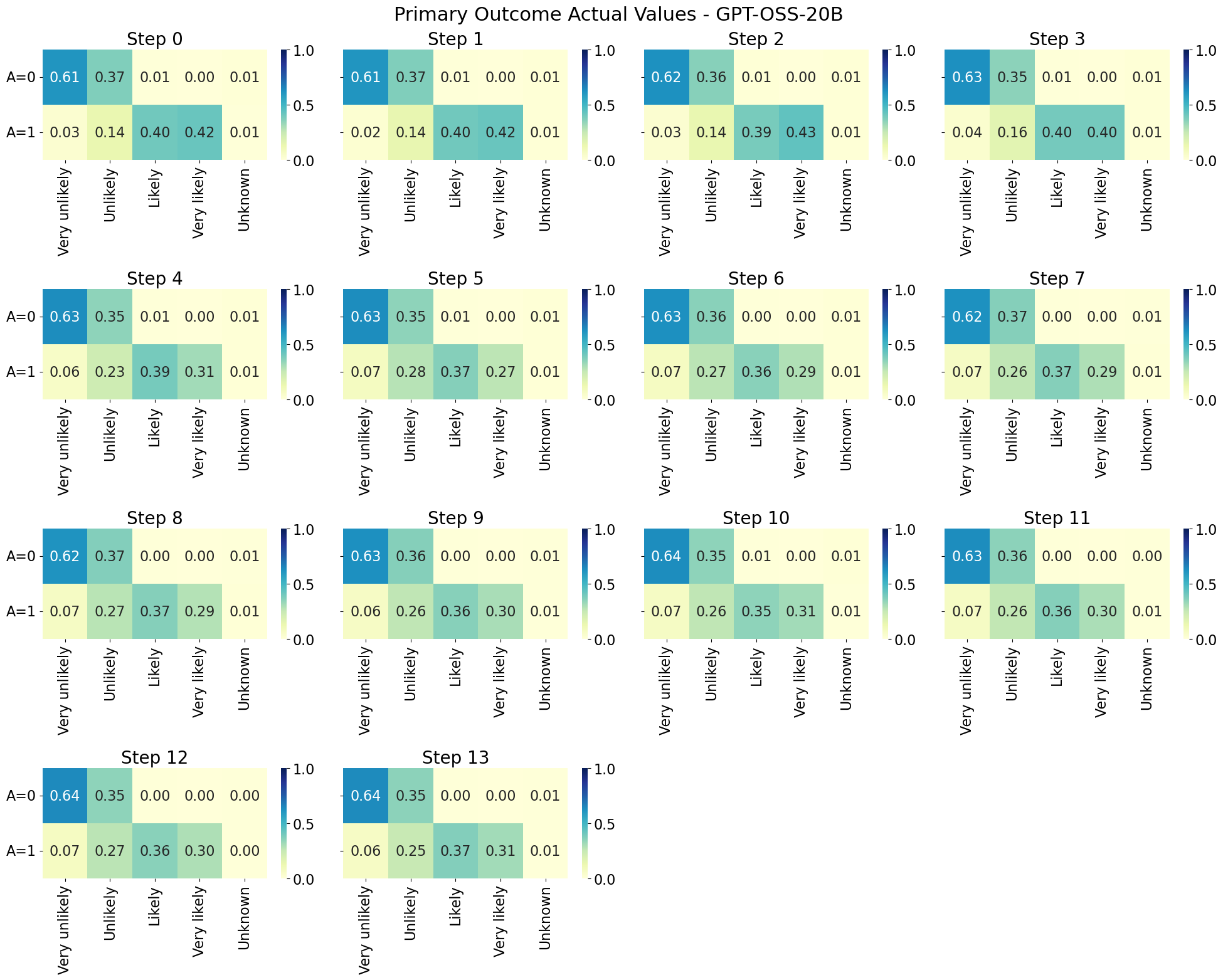}
    \caption{MovieLens (GPT-OSS-20B)}
    \label{fig:outcome_movielens_gpt-instruct}
  \end{subfigure}
  \caption{Primary outcome distributions faceted by intervention condition at adjustment iteration 0.}
  \label{fig:outcomes-step0}
\end{figure*}

\begin{figure}[h]
    \centering
    \includegraphics[width=\linewidth]{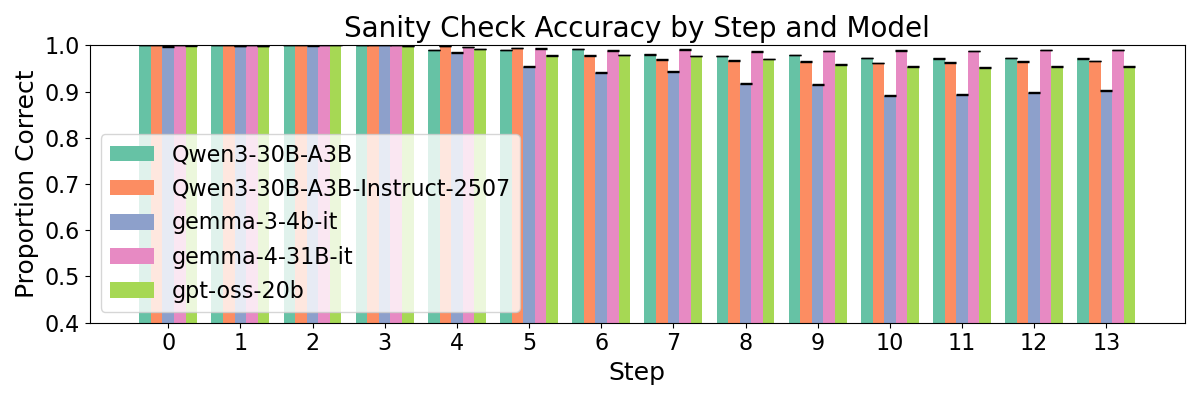}
    \caption{Proportion of specified persona attributes correctly reported after intervention dialogue in Book Opinions (30 personas with 10 trials each).}
    \label{fig:sanity_checks}
\end{figure}

% \textcolor{red}{[TODO: Restructure]}

% \begin{figure}[t]
%     \centering
%     \includegraphics[width=\linewidth]{figures/opinionqa-100p-30t/google-gemma-4-31B-it_nc_demo_marginal.png}
%     \caption{Marginal negative control outcome distributions over adjustment iterations (Gemma-4-31B, OpinionQA)}
%     \label{fig:ncdemo_marginal_gemma-4-it}
% \end{figure}

We first sanity check that the intervention has had the desired effect on the outcome. Across all three datasets, we observe that each intervention condition indeed induces the expected change in the outcome distribution (Figure \ref{fig:outcomes-step0}).

\textbf{Are user drift and responsiveness to confounder adjustment linked to model capability/capacity?} \; 
We explore whether models that are better at following instructions and maintaining consistent personas are both (a) more susceptible to user drift and (b) more responsive to adjustment. We attempt to proxy this persona-maintaining capability by checking how well models retain the information provided in the persona specification after the dialogue between the simulated user and the recommender LLM agent has been completed. After the dialogue, the simulated user is prompted again for the attributes that were specified in its initial persona prompt. In Figure \ref{fig:sanity_checks}, we plot the proportion of attributes that were correctly returned after the dialogue in the Book Opinions dataset. Because the list of attributes increases with each iteration, it is expected that the proportion correctly retained will decrease over iterations.

In this evaluation, Gemma-4-31B shows the best retention abilities and also is the model that seems most responsive to adjustment in the agent evaluation setting. Conversely, Gemma-3-4B and GPT-OSS-20B have the worst retention abilities and are also the two models to exhibit very little TVD even at iteration 0 in the MovieLens setting. This may provide evidence for our point that good models experience more user drift due to their strong abductive capabilities. Interestingly, while Gemma-3-4B has by far the worst retention abilities---which we expect due to its small size of only 4B---it is also more responsive to adjustment in the Book Opinions setting than the much larger Qwen3 models.
% Literature such as \citet{Tosato_Helbling_Mantilla-Ramos_Hegazy_Tosato_Lemay_Rish_Dumas_2026} could explain this phenomenon, i.e., potentially Gemma-3 as a model has good persona-simulating ability and the lack of information retention is due to the small size of the 4B model. \textcolor{red}{[TODO: Confirm]}

\textbf{Does the overall user population drift due to confounding adjustment?} \; 
Because we sample confounders to include in our persona specification, and conditioning on LLMs by prompting is not exact, the population we start with is potentially not the population we end with. That is, we stabilize the population \textit{across} interventions, but the overall population may drift over time. We note that this is an \textit{external validity} problem similar to those we mention in Section \ref{sec:related-work} and is not the focus of our work.

\begin{figure}[!h]
    \centering
    \includegraphics[width=\linewidth]{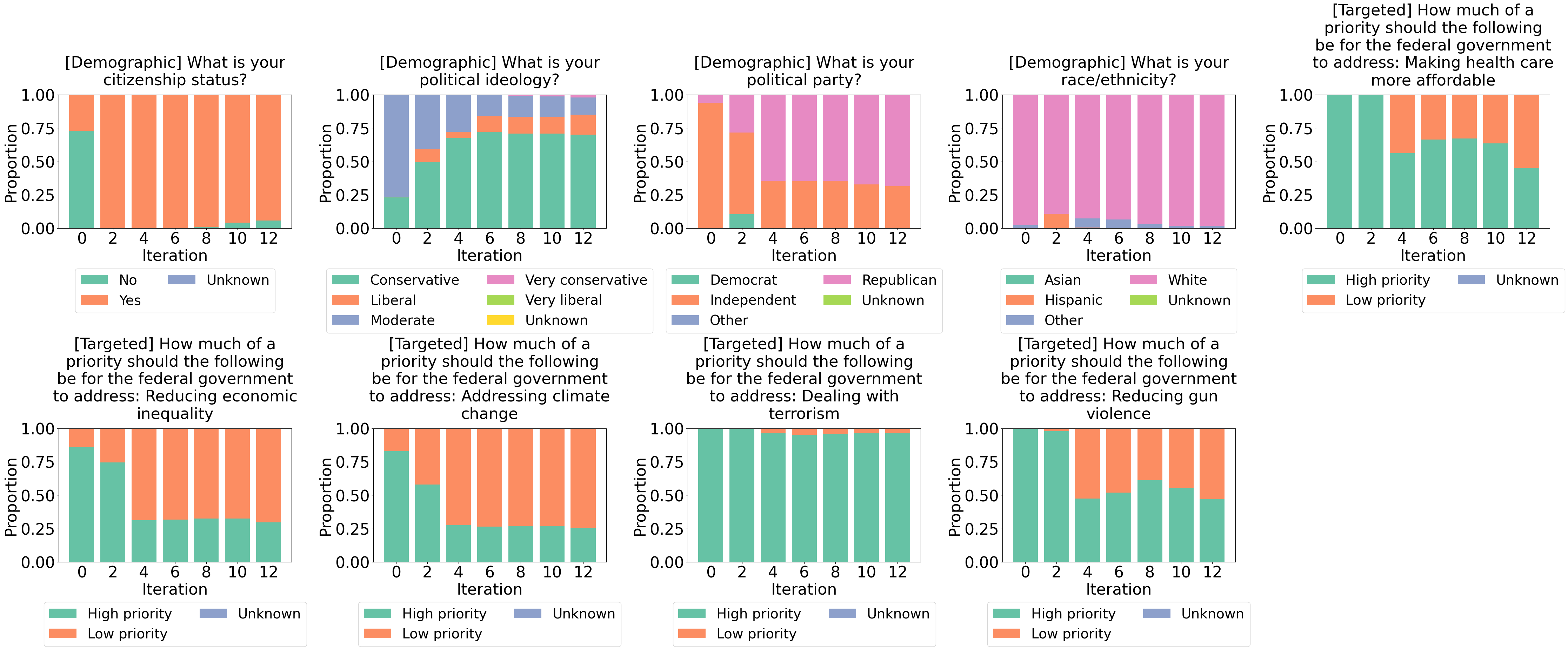}
    \caption{Marginal negative control outcome distributions over iterations of Gemma-4-31B on OpinionQA.}
    \label{fig:ncdemo_marginal_gemma-4-it}
\end{figure}

However, for the sake of analysis, we can check this directly by looking at the \textit{marginal} distribution of negative control outcomes $P(Z)$ over adjustment iterations, e.g. for instruction-tuned Gemma-4-31B on OpinionQA (Figure \ref{fig:ncdemo_marginal_gemma-4-it}). We see that some of these distributions do change over time, while others remain more or less constant.
% they are mostly minor (seemingly shifts of less than 0.1-0.15 probability). The exception is the \textit{citizenship status} negative control, wherein the respondent becomes much more likely to say that they are a citizen. This seems to be in response to the elicitation of the geographic region confounder, where the respondent is prompted to answer which region they are from and the candidate options are regions of the United States. (This is a bit of an odd phenomenon because the question does not ask if the respondent is a \textit{US} citizen, so it seems that some assumption is also being made on the model's part there.) In general, it is not unexpected that the shift is generally minimal since the confounder distributions are being elicited from the natural confounder distribution given the persona specification.
If significant drifts in the marginal population occur over adjustment iterations, then a possibility is to draw on prior approaches for improving external validity between iterations, such as the doubly robust approach proposed by \citet{guerdan2026doublyrobust}.

% If for some reason there is a significant drift in the marginal population between iteration steps, then a possibility is to use standard reweighting methods between the populations at each iteration to recover the effect in the original population.

% \subsection{Supplemental plots}

% \begin{figure}[!h]
%   \centering
%   % \begin{subfigure}[t]{0.33\linewidth}
%     \begin{subfigure}[t]{0.49\linewidth}
%     \centering
%     \includegraphics[width=\linewidth,trim=0cm 51.5cm 60.5cm 7cm,clip]{figures/movielens-100p-30t/openai-gpt-oss-20b_nc_demo_marginal.png}
%     \caption{GPT-OSS-20B}
%     \label{fig:ncdemomarginal_movielens_gptoss}
%   \end{subfigure}\hfill
%   % \begin{subfigure}[t]{0.33\linewidth}
%     \begin{subfigure}[t]{0.49\linewidth}
%     \centering
%     \includegraphics[width=\linewidth,trim=0cm 51.5cm 60.5cm 7cm,clip]{figures/movielens-100p-30t/google-gemma-4-31B-it_nc_demo_marginal.png}
%     \caption{Gemma-4-31B-it}
%     \label{fig:ncdemomarginal_movielens_gemma4}
%   \end{subfigure}
%   \caption{Marginal demographics-based negative control outcome distributions in MovieLens.}
% \end{figure}

\begin{figure}[!h]
  \centering
  % \begin{subfigure}[t]{0.33\linewidth}
    \begin{subfigure}[t]{0.4\linewidth}
    \centering
    \includegraphics[width=\linewidth,trim=0cm 0cm 90cm 0cm,clip]{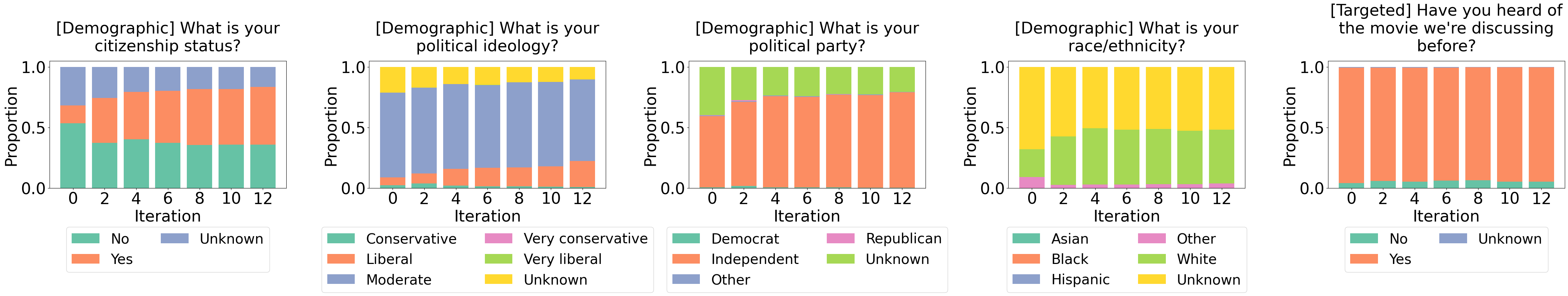}
    \caption{GPT-OSS-20B}
    \label{fig:ncdemomarginal_movielens_gptoss}
  \end{subfigure}\hfill
  % \begin{subfigure}[t]{0.33\linewidth}
    \begin{subfigure}[t]{0.4\linewidth}
    \centering
    \includegraphics[width=\linewidth,trim=0cm 0cm 90cm 0cm,clip]{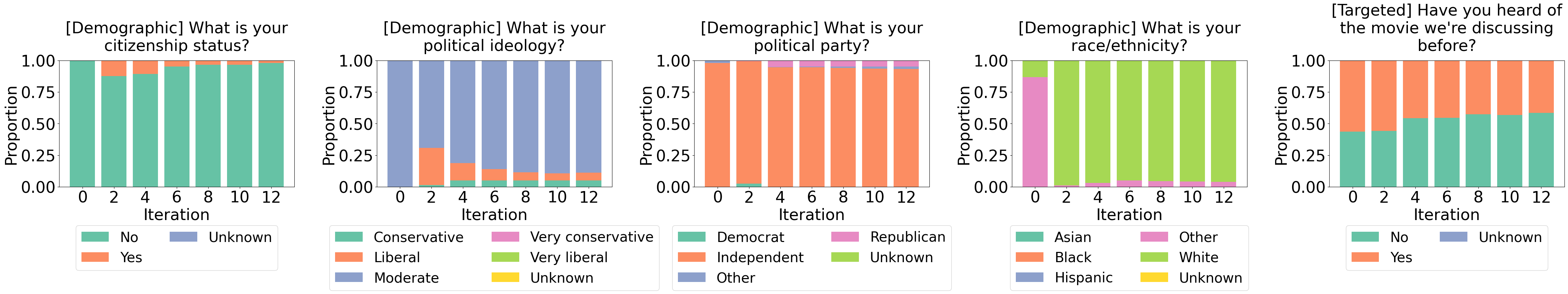}
    \caption{Gemma-4-31B-it}
    \label{fig:ncdemomarginal_movielens_gemma4}
  \end{subfigure}
  \caption{Marginal demographics-based negative control outcome distributions in MovieLens.}
\end{figure}

% \begin{figure}[!h]
%     \centering
%     \includegraphics[width=\linewidth]{figures/opinionqa-100p-30t/google-gemma-4-31B-it_nc_demo_marginal.png}
%     \caption{Marginal demographics-based negative control outcome distributions over iterations of Gemma-4-31B on OpinionQA.}
%     \label{fig:ncdemo_marginal_gemma-4-it}
% \end{figure}

\newpage
\section{Model details}
\label{sec:appendix:model_details}
\begin{table}[!h]
    \centering
    \caption{Technical details for language model implementations.}
    \vskip 0.1in
    \begin{tabular}{cccccc}
    \toprule
         Library & Model & Temperature & Top-$P$ & Top-$K$ \\
    \midrule
        \Verb|vllm| & \Verb|openai/gpt-oss-20b| & 1.0 & 1.0 & 0 \\
        \Verb|vllm| & \Verb|google/gemma-3-4b-it| & 1.0 & 0.95 & 64 \\
        \Verb|vllm| & \Verb|google/gemma-4-31B-it| & 1.0 & 0.95 & 64 \\
        \Verb|vllm| & \Verb|Qwen/Qwen3-30B-A3B| & 0.7 & 0.8 & 20 \\
        \Verb|vllm| & \Verb|Qwen/Qwen3-30B-A3B-Instruct-2507| & 0.7 & 0.8 & 20 \\
        Gemini API & \Verb|gemini-3-flash-preview| & 1.0 & 0.95 & 64 \\
    \bottomrule
    \end{tabular}
    \label{tab:language_reps}
\end{table}

To implement our simulations, we use several open-source large language models as detailed in Table \ref{tab:language_reps}. We evaluate \Verb|openai/gpt-oss-20b|, \Verb|Qwen/Qwen3-30B-A3B|, \Verb|Qwen/Qwen3-30B-A3B-Instruct-2507|, \Verb|google/gemma-4-31B-it|, and \Verb|google/gemma-3-4b-it| as user agents in our simulation. Additionally, \Verb|google/gemma-4-31B-it| is employed as the dialogue agent for book/movie discussions.

Finally, Gemini 3 Flash is accessed via the Gemini API and used as a user agent. Additional technical details are provided in Table \ref{tab:language_reps}.

\section{Dataset details}
\label{sec:appendix:dataset_details}
\begin{table}[h]
\centering
\caption{Overview of datasets}
\renewcommand{\arraystretch}{1.3}
\begin{tabularx}{\linewidth}{%
  >{\hsize=0.5\hsize\raggedright\arraybackslash}X
  >{\hsize=0.5\hsize\raggedright\arraybackslash}X
  >{\hsize=1.0\hsize\raggedright\arraybackslash}X%
}
\hline
\multicolumn{1}{c}{Dataset} & \multicolumn{1}{c}{License} & \multicolumn{1}{c}{URL} \\ \hline
OpinionQA \citep{pmlr-v202-santurkar23a} & Custom (Pew Research Non-Commercial) & \url{https://github.com/tatsu-lab/opinions_qa} \\
Book Opinions \citep{meister-etal-2025-benchmarking} & Custom / MIT & \url{https://github.com/nicolemeister/benchmarking-distributional-alignment} \\
MovieLens \citep{harper-2015-movielens} & GroupLens License (Non-Commercial) & \url{https://grouplens.org/datasets/movielens/} \\ \hline
\end{tabularx}
\label{tab:datasets}
\end{table}

\section{Computational resources}
\label{sec:appendix:compute}
Each experimental setting (e.g., one full run of a setting with $n$ personas $\times$ $T$ trials $\times$ 2 interventions) was executed on a single node with 8\,$\times$\,NVIDIA A100 GPUs. Across settings, we used a multi-node cluster of such machines (running settings and sweeps in parallel). Over all reported runs, we estimate total usage of approximately 3000 GPU-hours.

\section{Shared implementation details}
\label{sec:appendix:shared_implementation_details}
\paragraph{Negative control outcomes (demographic).}\mbox{}
\begin{lstlisting}
{
    "citizenship status": ["Yes", "No"],
    "political party": ["Republican", "Democrat", "Independent", "Other"],
    "political ideology": [
        "Very conservative",
        "Conservative",
        "Moderate",
        "Liberal",
        "Very liberal",
    ],
    "race": ["White", "Black", "Asian", "Hispanic", "Other"]
}
\end{lstlisting}

\paragraph{Confounders (demographic).}\mbox{}
\begin{lstlisting}
{
    "income": [
        "Less than $30,000",
        "$30,000-$50,000",
        "$50,000-$75,000",
        "$75,000-$100,000",
        "$100,000 or more",
    ],
    "census region": ["Northeast", "Midwest", "South", "West"],
    "education level": [
        "Less than high school",
        "High school graduate",
        "Some college, no degree",
        "Associate's degree",
        "College graduate/some postgrad",
        "Postgraduate",
    ],
    "marital status": [
        "Married",
        "Divorced",
        "Separated",
        "Widowed",
        "Never been married",
    ],
    "religion": [
        "Protestant",
        "Roman Catholic",
        "Mormon",
        "Orthodox",
        "Jewish",
        "Muslim",
        "Buddhist",
        "Hindu",
        "Atheist",
        "Agnostic",
        "Other",
        "Nothing in particular",
    ],
    "religious attendance frequency": [
        "More than once a week",
        "Once a week",
        "Once or twice a month",
        "A few times a year",
        "Seldom",
        "Never",
    ]
}
\end{lstlisting}

\section{Simulated user survey implementation details}
\label{sec:appendix:opinionqa_details}
\subsection{Simulated user survey prompts}

\paragraph{Synthetic user instantiation.}\mbox{}

\begin{lstlisting}
[SYSTEM]
You are acting as the following person: Persona:
Age: <AGE>
Sex: <SEX>
[Optional: Additional information about you:
<Confounder question from previous step> <Answer from previous step>
...]
\end{lstlisting}

\paragraph{Intervention (leading statements).}\mbox{}

\begin{lstlisting}
{
    "High priority": "Citing concerns over national security, strain on public resources, and the rule of law, many believe border security should be a top concern.",
    "Low priority": "Many advocates point out that immigrants contribute significantly to the economy and fill essential jobs, suggesting enforcement should be a lower priority than reform."
}
\end{lstlisting}

\paragraph{Primary outcome elicitation.}\mbox{}

\begin{lstlisting}
[SYSTEM]
<Persona description as above>

[USER]
<Leading statement 1 / Leading statement 2>

Question: How much of a priority should the following be for the federal government to address: Reducing illegal immigration
Please answer ONLY with one of the following options: High priority, Low priority
\end{lstlisting}

% \paragraph{Negative control outcome elicitation.}\mbox{}

% \begin{lstlisting}
% [SYSTEM]
% <Persona description as above>

% [USER]
% <Leading statement 1 / Leading statement 2>

% Question: <Control question>
% Options: <Option 1>, <Option 2>, ...
% Please answer ONLY with one of the options.
% \end{lstlisting}

\paragraph{Negative control / confounder elicitation (demographic attributes).}\mbox{}

\begin{lstlisting}
[SYSTEM]
<Persona description as above>

[USER]
<Leading statement 1 / Leading statement 2>

[USER]
Question: What is your <characteristic>?
Options: <Option 1>, <Option 2>, ...
Please answer ONLY with one of the options.
\end{lstlisting}

\paragraph{Negative control / confounder elicitation (targeted questions).}\mbox{}

\begin{lstlisting}
[SYSTEM]
<Persona description as above>

[USER]
<Leading statement 1 / Leading statement 2>

Question: <Negative control / confounder question>
Please answer in a brief sentence.
\end{lstlisting}

\subsection{Questions}

\paragraph{Negative control questions (targeted).}\mbox{}
\begin{lstlisting}
{
    "question": "How much of a priority should the following be for the federal government to address: Making health care more affordable",
    "options": [
      "High priority",
      "Low priority"
    ]
},
{
    "question": "How much of a priority should the following be for the federal government to address: Reducing economic inequality",
    "options": [
      "High priority",
      "Low priority"
    ]
},
{
    "question": "How much of a priority should the following be for the federal government to address: Addressing climate change",
    "options": [
      "High priority",
      "Low priority"
    ]
},
{
    "question": "How much of a priority should the following be for the federal government to address: Dealing with terrorism",
    "options": [
      "High priority",
      "Low priority"
    ],
},
{
    "question": "How much of a priority should the following be for the federal government to address: Reducing gun violence",
    "options_condensed": [
      "High priority",
      "Low priority"
    ]
}
\end{lstlisting}

\paragraph{Confounder questions (targeted).}\mbox{}
\begin{lstlisting}
{
  "Political & Ideological Affiliation": [
    "How important is party loyalty to you when forming opinions on policy?",
    "How do you view the role of the federal government in daily life?"
  ],
  "Economic Beliefs & Self-Interest": [
    "Do you work in an industry that frequently employs immigrant labor (e.g., agriculture, construction, hospitality)?",
    "How do you perceive the impact of immigration on your local economy?"
  ],
  "National Security & Law Enforcement Views": [
    "How often do you worry about crime in your neighborhood?",
    "Do you believe the US borders are currently secure?"
  ],
  "Humanitarian & Ethical Considerations": [
    "How much do you value empathy in public policy?",
    "How often do you donate to or volunteer for charitable causes?"
  ],
  "Cultural & Social Identity": [
    "How often do you interact with people from different cultural backgrounds?",
    "How do you feel when you hear languages other than English spoken in public?"
  ],
  "Information & Media Consumption": [
    "Which cable news network do you watch most frequently?",
    "Do you fact-check information you see on social media?"
  ],
  "Personal Experience & Proximity": [
    "Have you ever lived in a country other than the United States?",
    "Has your community seen a significant change in its immigrant population recently?"
  ],
  "Trust in Institutions": [
    "How much do you trust the federal government to do what is right?",
    "Do you believe your voice matters in government decisions?"
  ],
  "Views on Government Role & Scope": [
    "How much responsibility should the federal government have in regulating the economy?",
    "What is your view on government spending for social welfare programs?"
  ],
  "General Worldview & Personality Traits": [
    "How open are you to new experiences?",
    "How important is it to you to conform to social norms?"
  ]
}
\end{lstlisting}

\section{Assistive agent evaluation implementation details}
\label{sec:appendix:multiturn_agent_details}
\subsection{Selected books}
The 5 books randomly chosen for discussion in the evaluation are:
\begin{itemize}
    \item \textit{Way Out There In The Blue: Reagan, Star Wars And The End Of The Cold War} by Frances Fitzgerald
    \item \textit{There There} by Tommy Orange
    \item \textit{Redeployment} by Phil Klay
    \item \textit{The Door} by Magda Szabo
    \item \textit{Nothing To Be Frightened Of} by Julian Barnes
\end{itemize}

\subsection{Selected movies}
The 5 movies randomly chosen for discussion in the evaluation are:
\begin{itemize}
    \item \textit{Other Side of Sunday, The (Søndagsengler) (1996)} (Comedy|Drama)
    \item \textit{Death in Brunswick (1991)} (Comedy)
    \item \textit{Buddy (1997)} (Adventure|Children's|Drama)
    \item \textit{Paris, Texas (1984)} (Drama)
    \item \textit{Lovers of the Arctic Circle, The (Los Amantes del Círculo Polar) (1998)} (Drama|Romance)
\end{itemize}

\subsection{Prompts}

\paragraph{Book/movie agent instantiation (with intervention).}\mbox{}

\begin{lstlisting}
[SYSTEM]
You are a helpful <book / movie> recommender interacting with a user. Your goal is to help the reader choose a <book / movie>. Discuss the following <book / movie>: <<title> by <author> / <movie: <title>, genres: <genres>>.
<Discuss this <book / movie> in a positively-biased way. / Discuss this <book / movie> in a negatively-biased way.>

[USER]
I'm looking for a <book to read / movie to watch>. Can you help me find one?
\end{lstlisting}

\paragraph{Synthetic user instantiation.}\mbox{}

\begin{lstlisting}
[SYSTEM]
You are acting as the following person:
Age: <AGE>
Gender: <GENDER>
[Optional: Additional information about you:
<Question from previous step> <Answer from previous step>
...]
You are looking for a <book to read / movie to watch>.

[USER]
<Book/movie agent's response>
\end{lstlisting}

\paragraph{Primary outcome elicitation.}\mbox{}

\begin{lstlisting}
[SYSTEM]
<Synthetic user system prompt from above>

[USER]
<Book/movie agent message 1>

[ASSISTANT]
<Synthetic user message 1>

... (more turns) ...

[USER]
<Book/movie agent message 3>

[ASSISTANT]
<Synthetic agent message 3>

[USER]
How likely are you to <read the book / watch the movie> '<title>'? Answer with one of:
1: Very unlikely
2: Unlikely
3: Likely
4: Very likely
\end{lstlisting}

% \paragraph{Negative control outcome elicitation.}\mbox{}

% \begin{lstlisting}
% [SYSTEM]
% <Synthetic user system prompt from above>

% ... (Dialogue history: USER and ASSISTANT messages) ...

% [USER]
% Question: <Control question>
% Options: <Option 1>, <Option 2>, ... 
% Please answer ONLY with one of the options.
% \end{lstlisting}

\paragraph{Negative control / confounder elicitation (demographic attributes).}\mbox{}

\begin{lstlisting}
[SYSTEM]
<Synthetic user system prompt from above>

... (Dialogue history: USER and ASSISTANT messages) ...

[USER]
Question: What is your <characteristic>?
Options: <Option 1>, <Option 2>, ... 
Please answer ONLY with one of the options.
\end{lstlisting}

\paragraph{Negative control / confounder elicitation (targeted question).}\mbox{}

\begin{lstlisting}
[SYSTEM]
<Synthetic user system prompt from above>

... (Dialogue history: USER and ASSISTANT messages) ...

[USER]
Question: <Negative control / confounder question>
Options: <Option 1>, <Option 2>, ...
Please answer ONLY with one of the options.
\end{lstlisting}

\subsection{Questions}

\paragraph{Negative control questions (targeted).}\mbox{}
\begin{lstlisting}
{
    "question": "Have you heard of the <book / movie> we're discussing before?",
    "options": [
      "Yes",
      "No"
    ]
}
\end{lstlisting}

\paragraph{Confounder questions (targeted).}\mbox{}
\begin{lstlisting}
{
    "<Reading / Movie Watching> Habits & Logistics": [
      "What is your favorite <book / movie> genre?",
      "How many <books do you read / movies do you watch> in a year?"
    ],
    "<Book / Movie> Selection & Criticality": [
      "How do you typically find out about new <books / movies>?",
      "Do you read <book / movie> reviews before deciding to <read a book / watch a movie>?"
    ],
    "Lifestyle & Time Allocation": [
      "How much leisure time do you have per day?",
      "What is your main occupation?"
    ],
    "Media & Leisure Habits": [
      "What is your primary source of news?",
      "How often do you attend cultural events (e.g., museums, concerts)?"
    ],
    "Personality & Cognitive Style": [
      "How easily are you influenced by others' opinions?",
      "Do you consider yourself an empathetic person?"
    ],
    "Decision Making & Outlook": [
      "Do you consider yourself a creative person?",
      "Do you enjoy debating topics with others?"
    ],
    "Emotional Well-being & Mood": [
      "How would you describe your general mood today?",
      "How often do you feel bored in your free time?"
    ],
    "Social & Personal Fulfillment": [
      "Are you currently seeking new challenges in your life?",
      "How optimistic are you about the future?"
    ],
    "Personal Values (Part 1)": [
      "How important is personal freedom and independence to you?",
      "How important is social status to you?"
    ],
    "Personal Values (Part 2)": [
      "How important is maintaining traditions to you?",
      "What value do you place on diversity and inclusion?"
    ]
}
\end{lstlisting}

\newpage
\section{Additional plots}
\label{sec:appendix:additional_plots}
% \textcolor{red}{[TODO: Don't forget to uncomment]}
% Each figure shows stacked distributions: (a) negative control outcomes and (b) primary outcomes.

\subsection{OpinionQA}

\begin{figure*}[h]
  \centering
  \begin{subfigure}[t]{\textwidth}
    \centering
    \includegraphics[width=\linewidth]{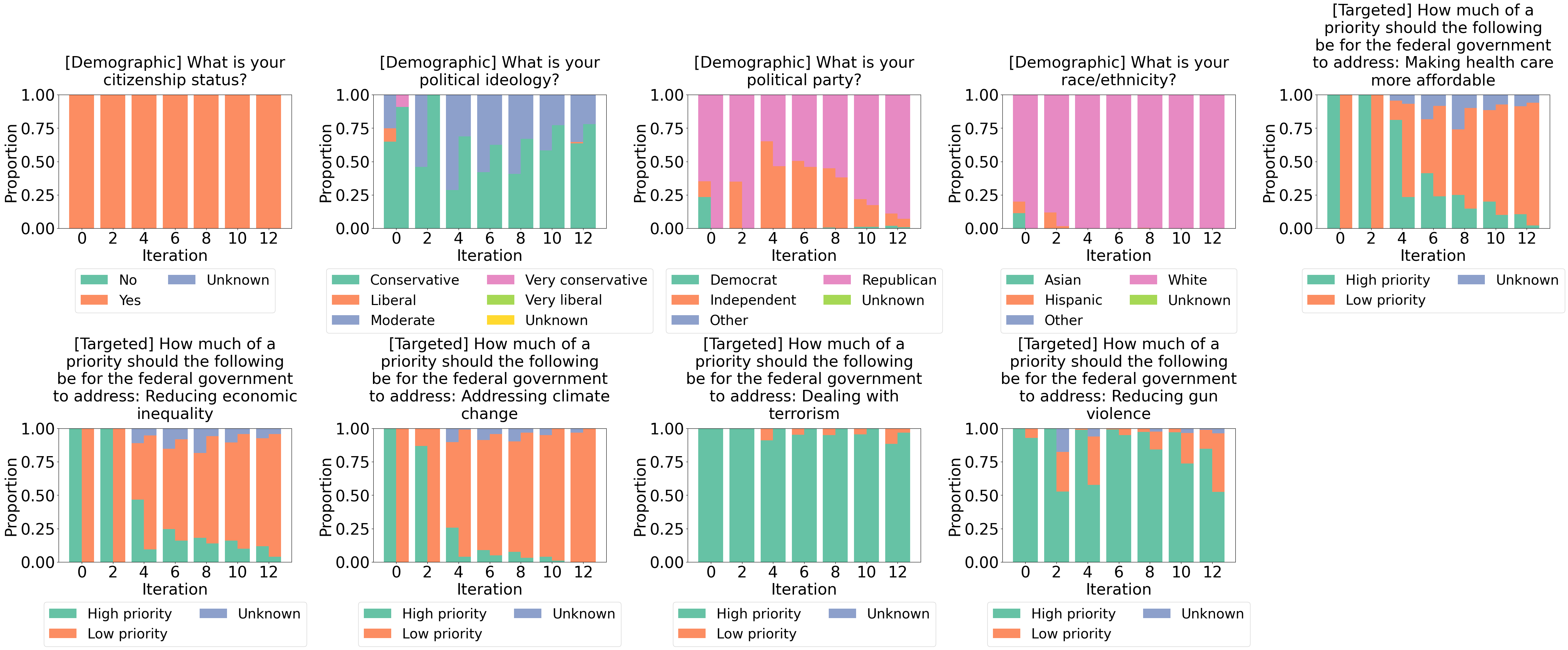}
    \caption{Negative control outcomes.}
  \end{subfigure}\\
  \begin{subfigure}[t]{0.49\textwidth}
    \centering
    \includegraphics[width=\linewidth]{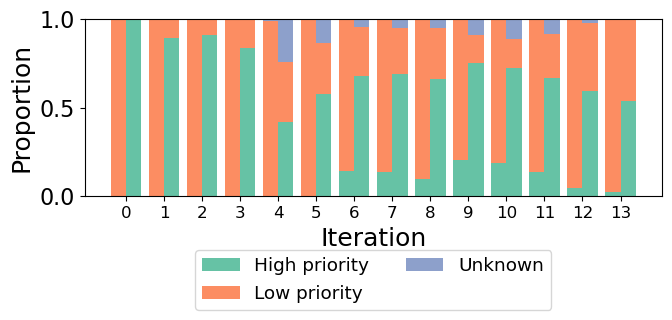}
    \caption{Primary outcomes.}
  \end{subfigure}
  \caption{OpinionQA outcome distributions (Gemma-3 4B-it). At each adjustment iteration, $A=0$ is on the left and $A=1$ is on the right.}
  \label{fig:app-opinionqa-gemma3-stacked}
\end{figure*}

\begin{figure*}[h]
  \centering
  \begin{subfigure}[t]{\textwidth}
    \centering
    \includegraphics[width=\linewidth]{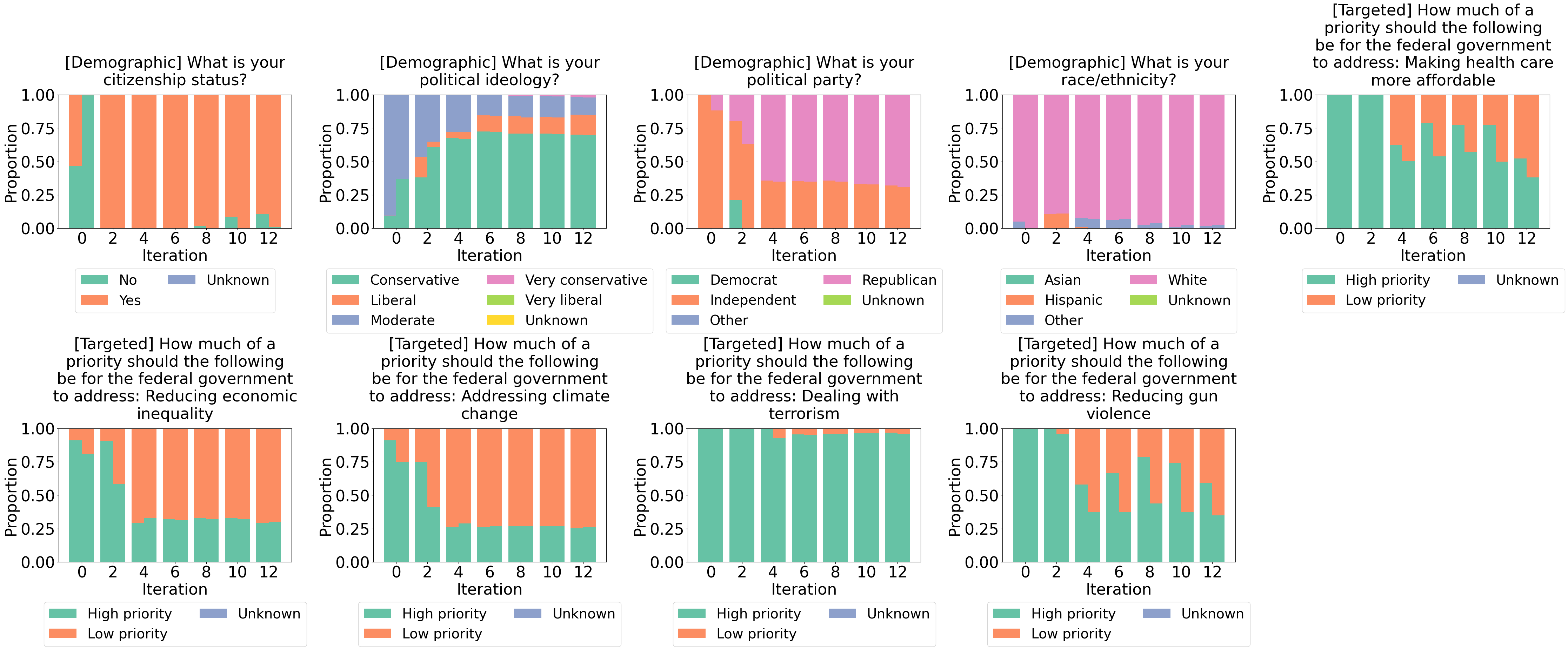}
    \caption{Negative control outcomes.}
  \end{subfigure}\\
  \begin{subfigure}[t]{0.49\textwidth}
    \centering
    \includegraphics[width=\linewidth]{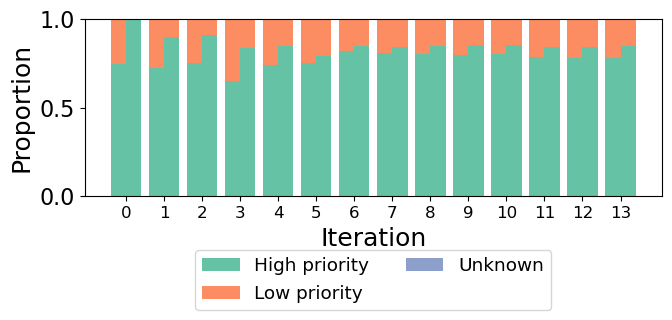}
    \caption{Primary outcomes.}
  \end{subfigure}
  \caption{OpinionQA outcome distributions (Gemma-4 31B-it). At each adjustment iteration, $A=0$ is on the left and $A=1$ is on the right.}
  \label{fig:app-opinionqa-gemma4-stacked}
\end{figure*}

\begin{figure*}[h]
  \centering
  \begin{subfigure}[t]{\textwidth}
    \centering
    \includegraphics[width=\linewidth]{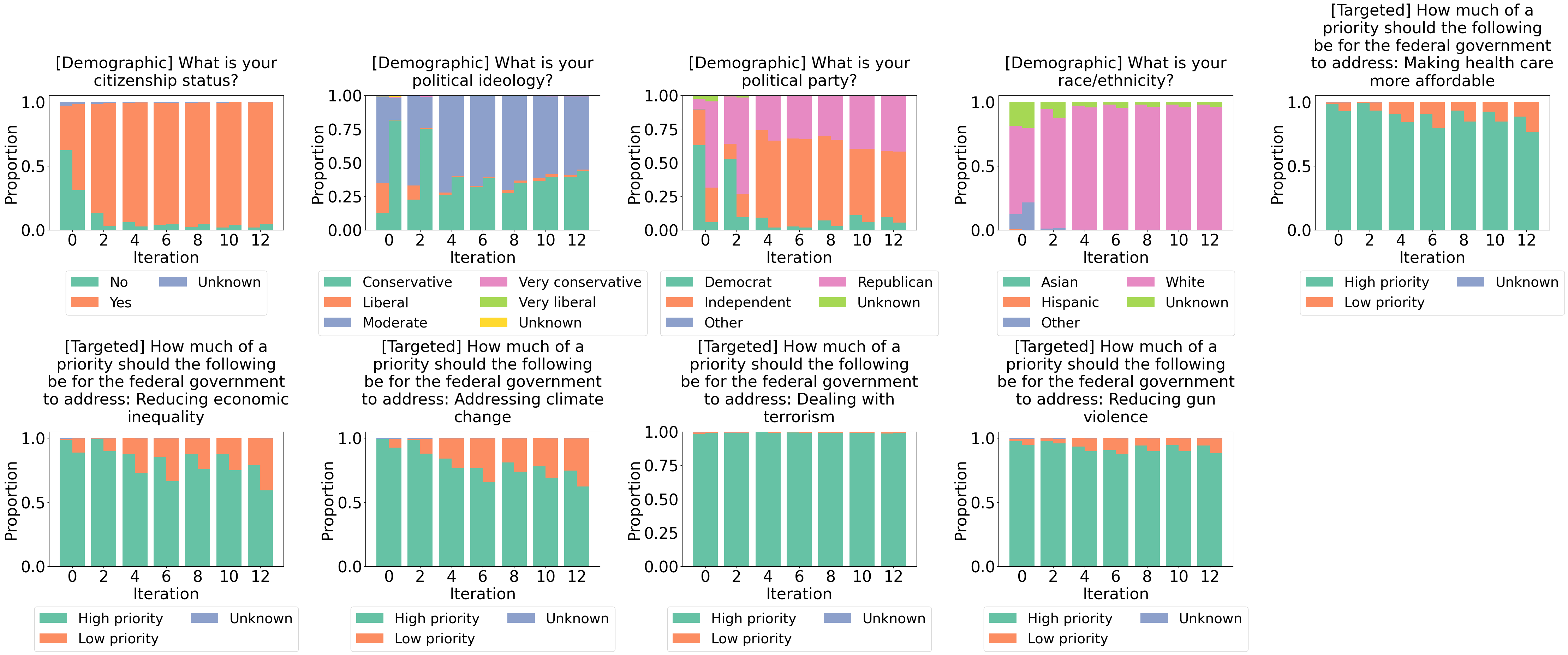}
    \caption{Negative control outcomes.}
  \end{subfigure}\\
  \begin{subfigure}[t]{0.49\textwidth}
    \centering
    \includegraphics[width=\linewidth]{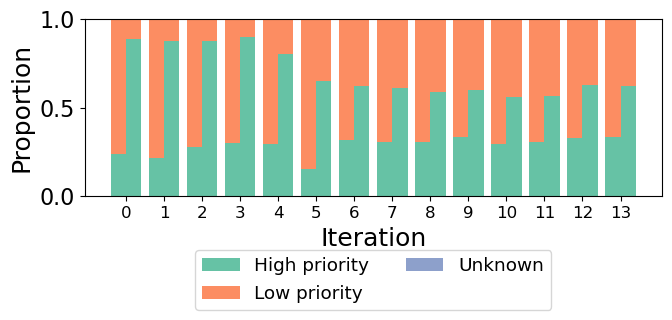}
    \caption{Primary outcomes.}
  \end{subfigure}
  \caption{OpinionQA outcome distributions (GPT-OSS 20B). At each adjustment iteration, $A=0$ is on the left and $A=1$ is on the right.}
  \label{fig:app-opinionqa-gptoss20b-stacked}
\end{figure*}

\begin{figure*}[h]
  \centering
  \begin{subfigure}[t]{\textwidth}
    \centering
    \includegraphics[width=\linewidth]{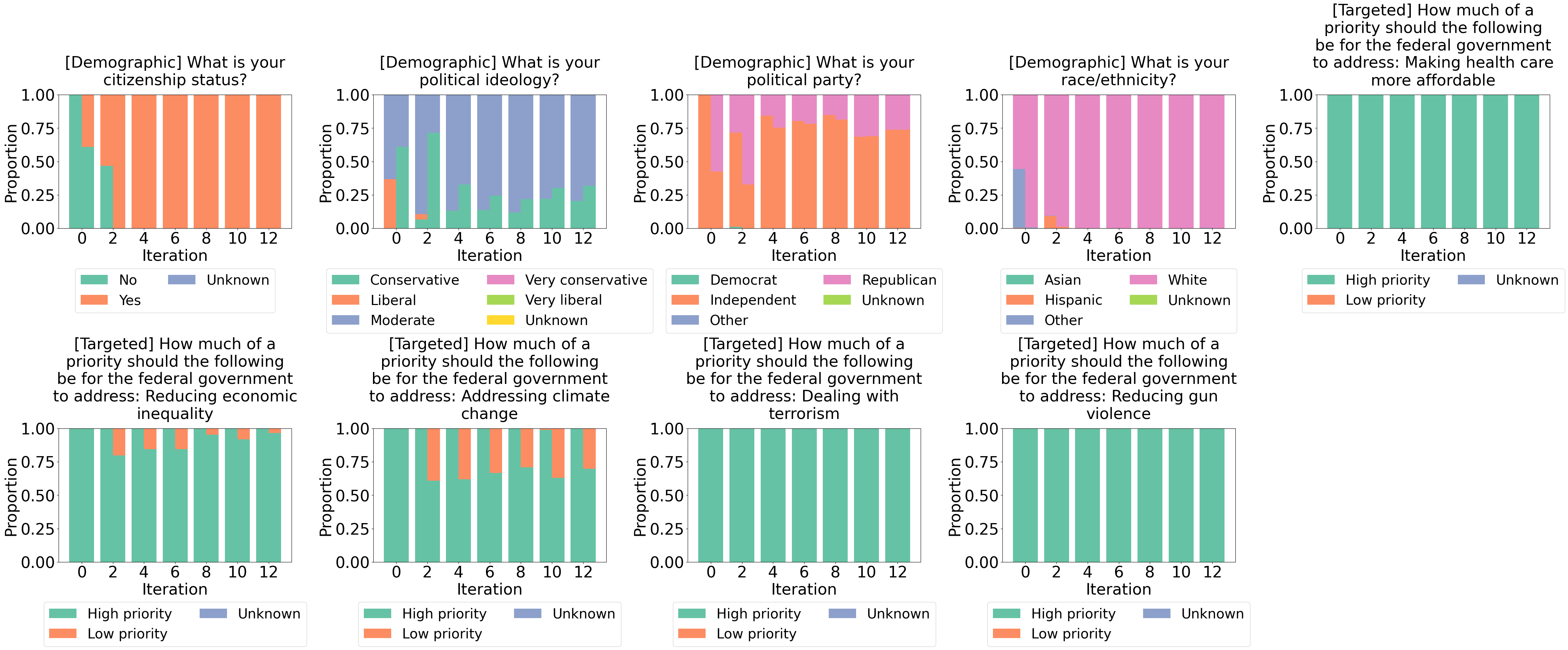}
    \caption{Negative control outcomes.}
  \end{subfigure}\\
  \begin{subfigure}[t]{0.49\textwidth}
    \centering
    \includegraphics[width=\linewidth]{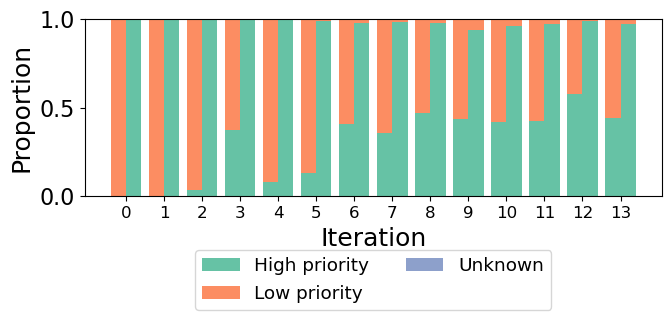}
    \caption{Primary outcomes.}
  \end{subfigure}
  \caption{OpinionQA outcome distributions (Qwen3 30B-A3B). At each adjustment iteration, $A=0$ is on the left and $A=1$ is on the right.}
  \label{fig:app-opinionqa-qwen-a3b-stacked}
\end{figure*}

\begin{figure*}[h]
  \centering
  \begin{subfigure}[t]{\textwidth}
    \centering
    \includegraphics[width=\linewidth]{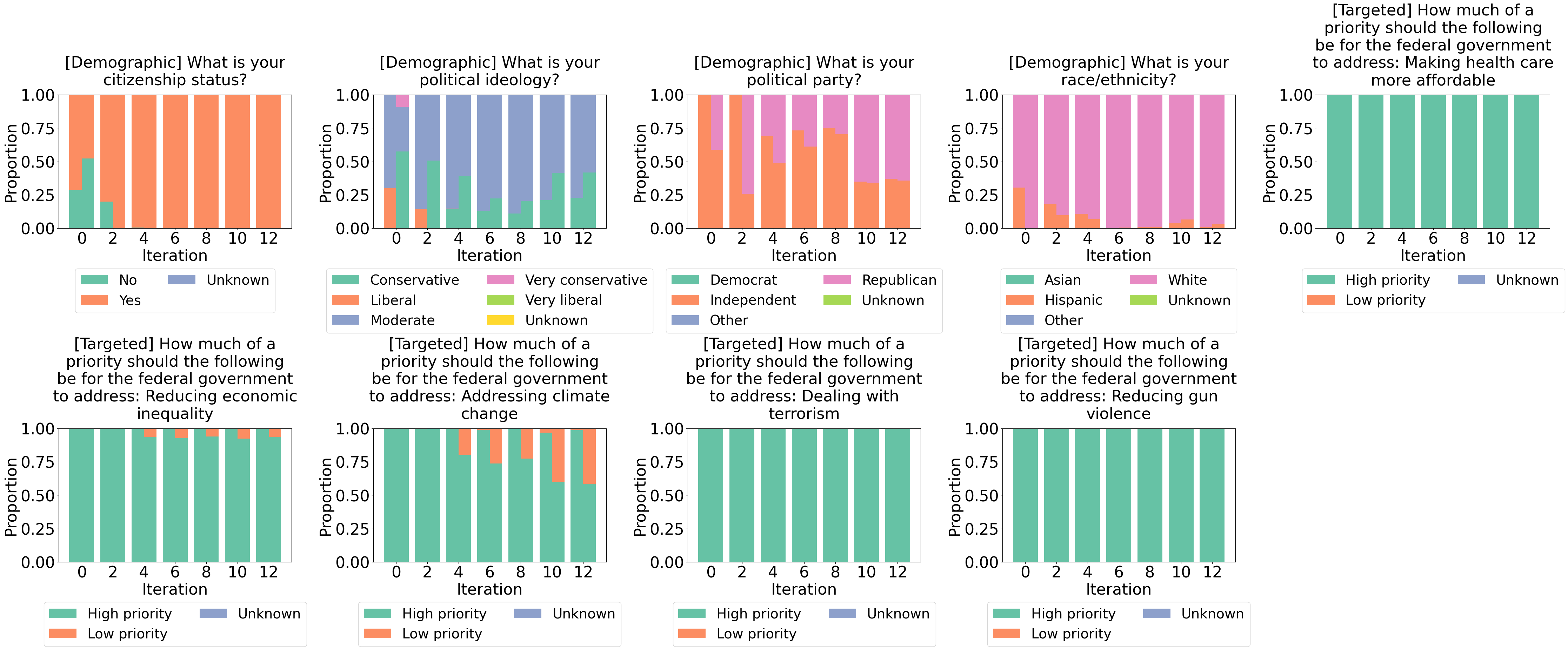}
    \caption{Negative control outcomes.}
  \end{subfigure}\\
  \begin{subfigure}[t]{0.49\textwidth}
    \centering
    \includegraphics[width=\linewidth]{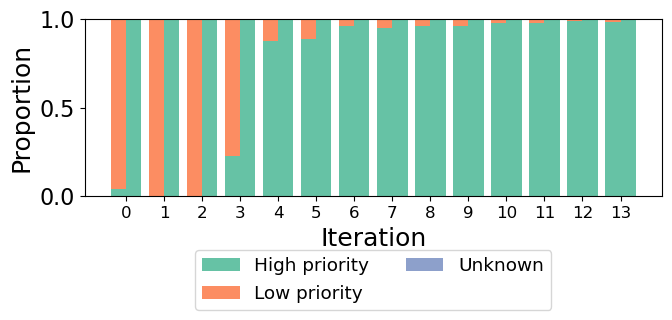}
    \caption{Primary outcomes.}
  \end{subfigure}
  \caption{OpinionQA outcome distributions (Qwen3 30B-A3B-Instruct-2507). At each adjustment iteration, $A=0$ is on the left and $A=1$ is on the right.}
  \label{fig:app-opinionqa-qwen-a3b-inst-stacked}
\end{figure*}

\begin{figure*}[h]
  \centering
  \begin{subfigure}[t]{\textwidth}
    \centering
    \includegraphics[width=\linewidth]{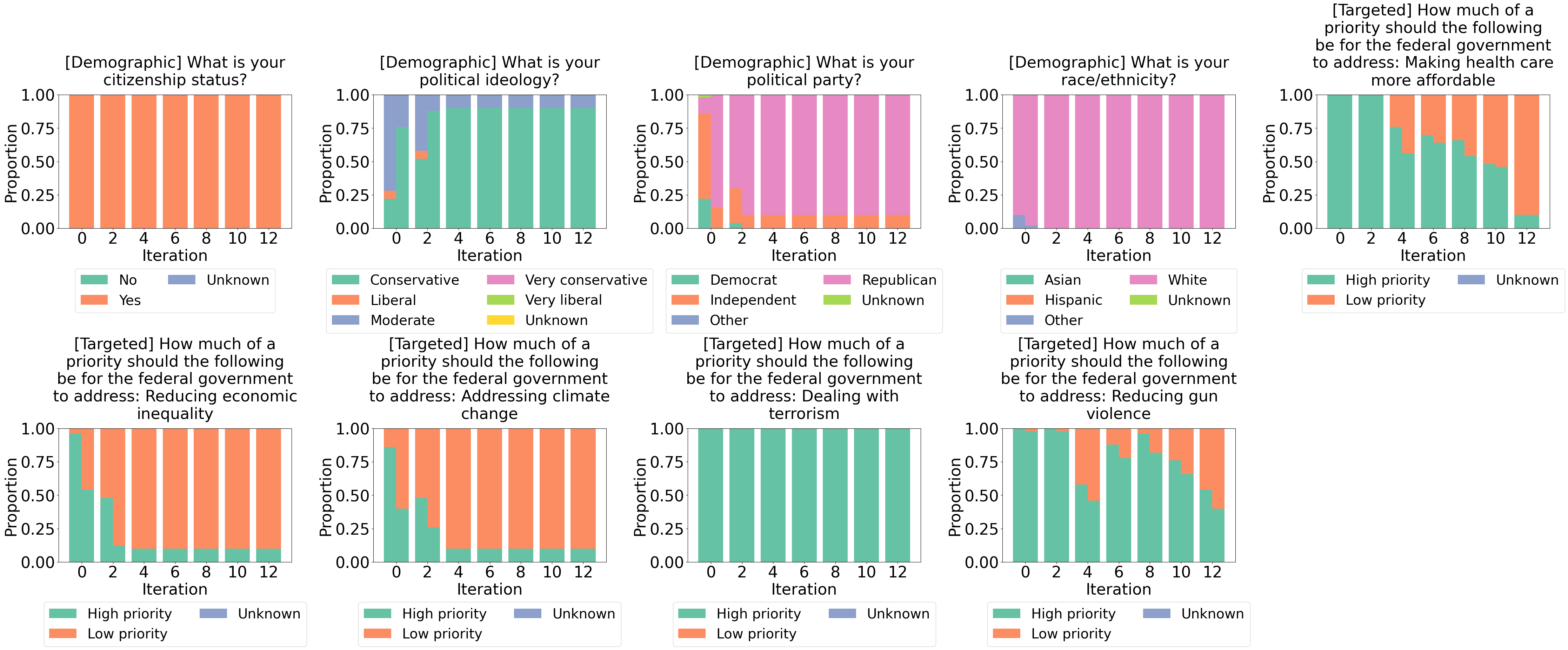}
    \caption{Negative control outcomes.}
  \end{subfigure}\\
  \begin{subfigure}[t]{0.49\textwidth}
    \centering
    \includegraphics[width=\linewidth]{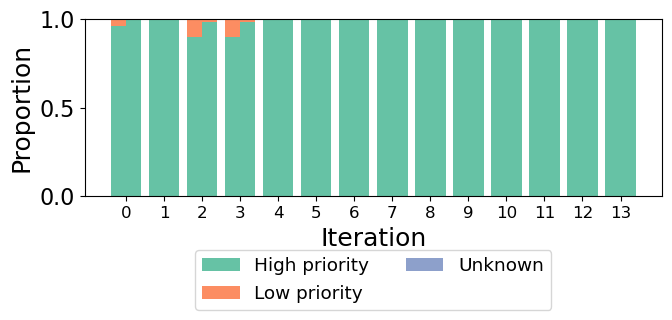}
    \caption{Primary outcomes.}
  \end{subfigure}
  \caption{OpinionQA outcome distributions (Gemini 3 Flash). At each adjustment iteration, $A=0$ is on the left and $A=1$ is on the right.}
  \label{fig:app-opinionqa-gemini3flash-stacked}
\end{figure*}

\clearpage
\subsection{Book Opinions}

\begin{figure*}[h]
  \centering
  \begin{subfigure}[t]{\textwidth}
    \centering
    \includegraphics[width=\linewidth]{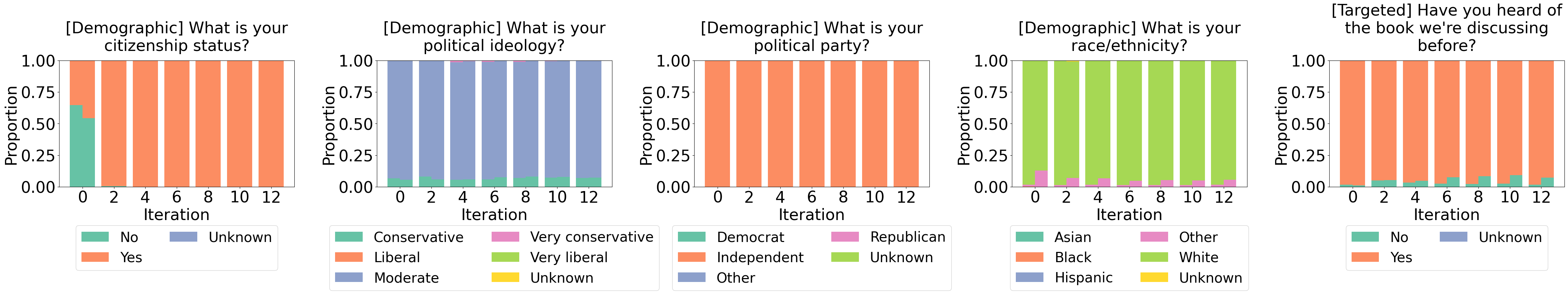}
    \caption{Negative control outcomes.}
  \end{subfigure}\\
  \begin{subfigure}[t]{0.49\textwidth}
    \centering
    \includegraphics[width=\linewidth]{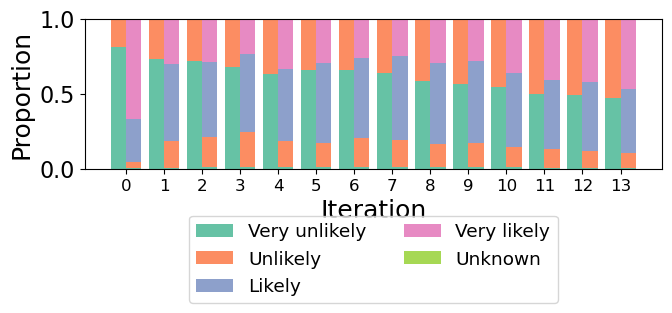}
    \caption{Primary outcomes.}
  \end{subfigure}
  \caption{Book Opinions outcome distributions (Gemma-3 4B-it). At each adjustment iteration, $A=0$ is on the left and $A=1$ is on the right.}
  \label{fig:app-nyt-gemma3-stacked}
\end{figure*}

\begin{figure*}[h]
  \centering
  \begin{subfigure}[t]{\textwidth}
    \centering
    \includegraphics[width=\linewidth]{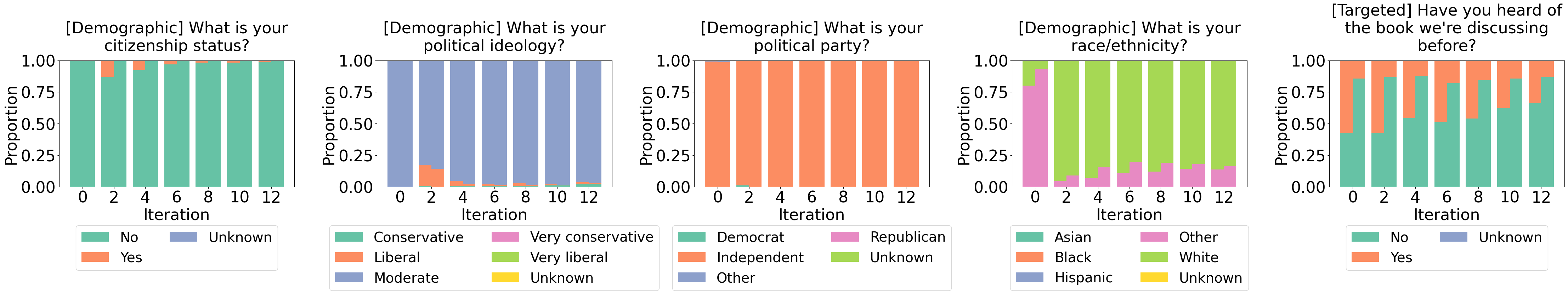}
    \caption{Negative control outcomes.}
  \end{subfigure}\\
  \begin{subfigure}[t]{0.49\textwidth}
    \centering
    \includegraphics[width=\linewidth]{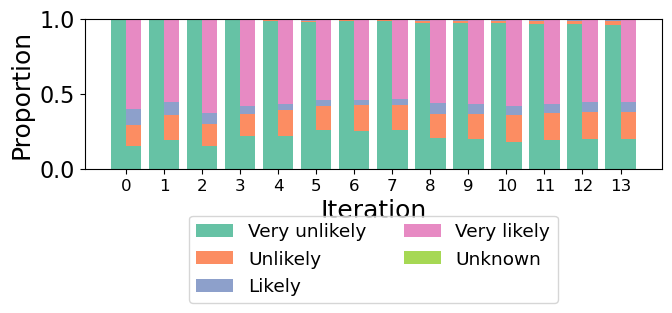}
    \caption{Primary outcomes.}
  \end{subfigure}
  \caption{Book Opinions outcome distributions (Gemma-4 31B-it). At each adjustment iteration, $A=0$ is on the left and $A=1$ is on the right.}
  \label{fig:app-nyt-gemma4-stacked}
\end{figure*}

\begin{figure*}[h]
  \centering
  \begin{subfigure}[t]{\textwidth}
    \centering
    \includegraphics[width=\linewidth]{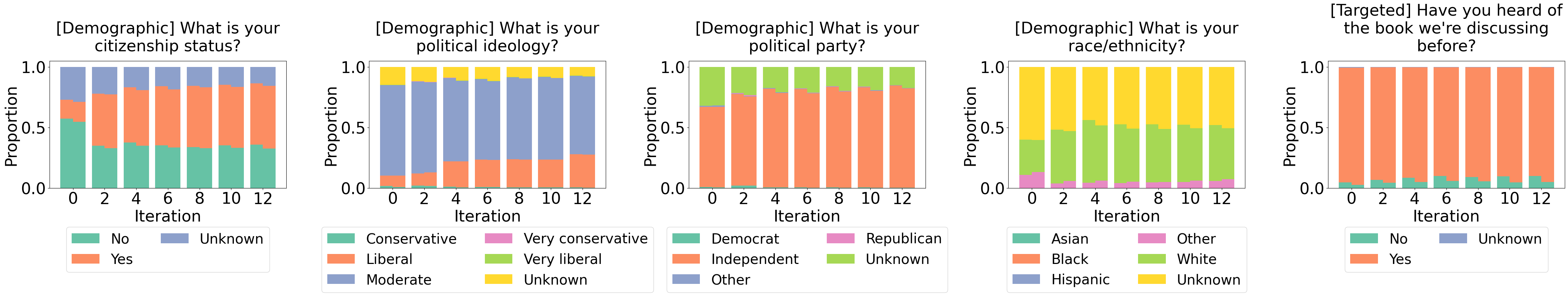}
    \caption{Negative control outcomes.}
  \end{subfigure}\\
  \begin{subfigure}[t]{0.49\textwidth}
    \centering
    \includegraphics[width=\linewidth]{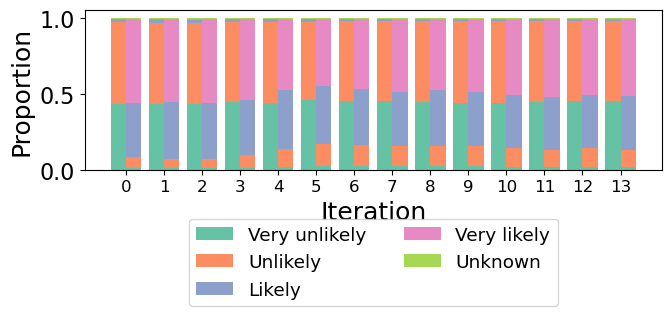}
    \caption{Primary outcomes.}
  \end{subfigure}
  \caption{Book Opinions outcome distributions (GPT-OSS 20B). At each adjustment iteration, $A=0$ is on the left and $A=1$ is on the right.}
  \label{fig:app-nyt-gptoss20b-stacked}
\end{figure*}

\begin{figure*}[h]
  \centering
  \begin{subfigure}[t]{\textwidth}
    \centering
    \includegraphics[width=\linewidth]{figures/nyt-100p-30t/Qwen-Qwen3-30B-A3B_nc_stacked.png}
    \caption{Negative control outcomes.}
  \end{subfigure}\\
  \begin{subfigure}[t]{0.49\textwidth}
    \centering
    \includegraphics[width=\linewidth]{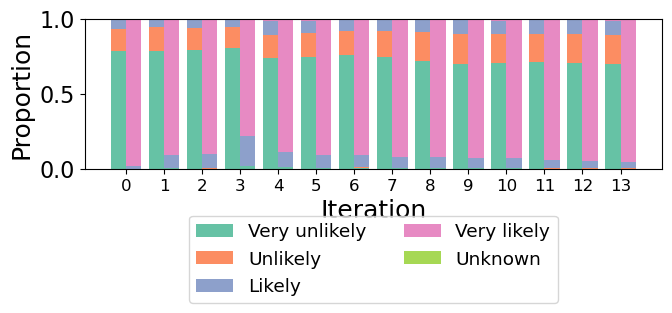}
    \caption{Primary outcomes.}
  \end{subfigure}
  \caption{Book Opinions outcome distributions (Qwen3 30B-A3B). At each adjustment iteration, $A=0$ is on the left and $A=1$ is on the right.}
  \label{fig:app-nyt-qwen-a3b-stacked}
\end{figure*}

\begin{figure*}[h]
  \centering
  \begin{subfigure}[t]{\textwidth}
    \centering
    \includegraphics[width=\linewidth]{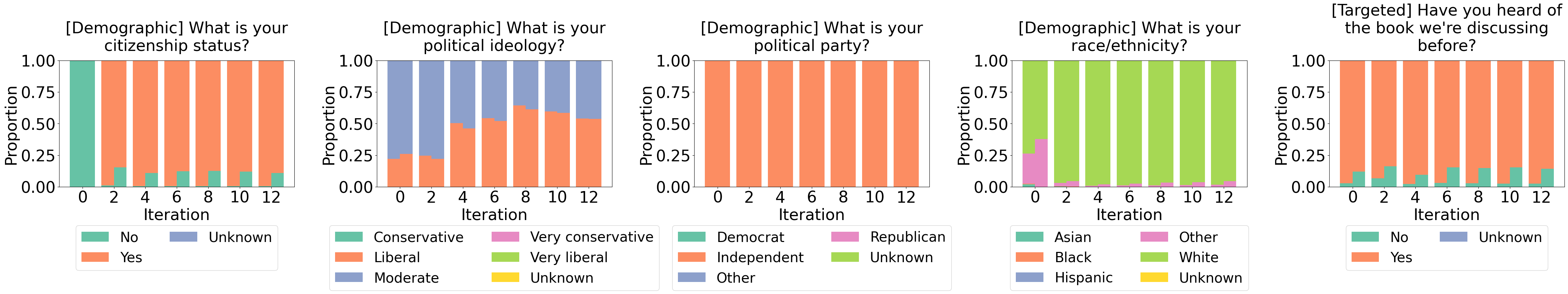}
    \caption{Negative control outcomes.}
  \end{subfigure}\\
  \begin{subfigure}[t]{0.49\textwidth}
    \centering
    \includegraphics[width=\linewidth]{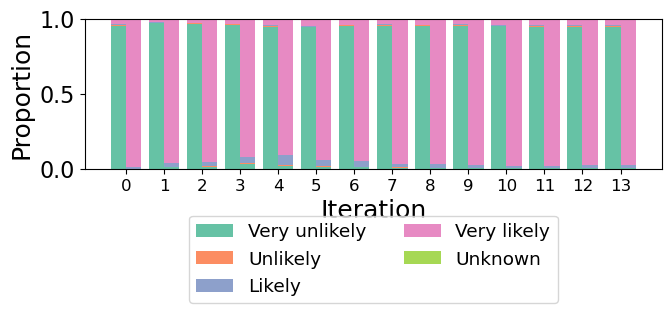}
    \caption{Primary outcomes.}
  \end{subfigure}
  \caption{Book Opinions outcome distributions (Qwen3 30B-A3B-Instruct-2507). At each adjustment iteration, $A=0$ is on the left and $A=1$ is on the right.}
  \label{fig:app-nyt-qwen-a3b-inst-stacked}
\end{figure*}

\begin{figure*}[h]
  \centering
  \begin{subfigure}[t]{\textwidth}
    \centering
    \includegraphics[width=\linewidth]{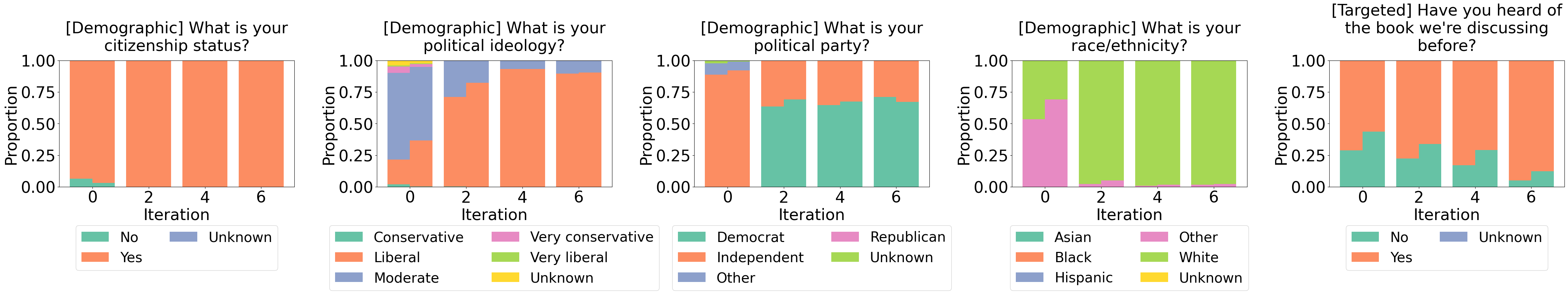}
    \caption{Negative control outcomes.}
  \end{subfigure}\\
  \begin{subfigure}[t]{0.49\textwidth}
    \centering
    \includegraphics[width=\linewidth]{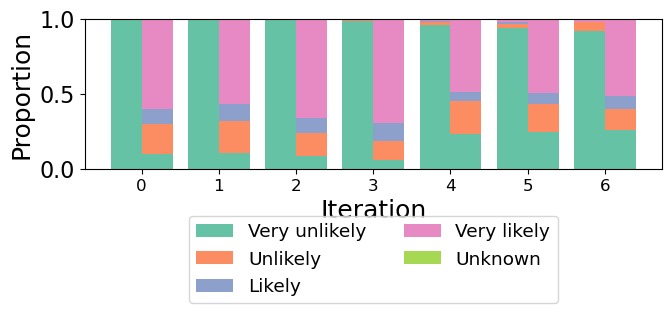}
    \caption{Primary outcomes.}
  \end{subfigure}
  \caption{Book Opinions outcome distributions (Gemini 3 Flash). At each adjustment iteration, $A=0$ is on the left and $A=1$ is on the right.}
  \label{fig:app-nyt-gemini3flash-stacked}
\end{figure*}

\clearpage
\subsection{MovieLens}

\begin{figure*}[h]
  \centering
  \begin{subfigure}[t]{\textwidth}
    \centering
    \includegraphics[width=\linewidth]{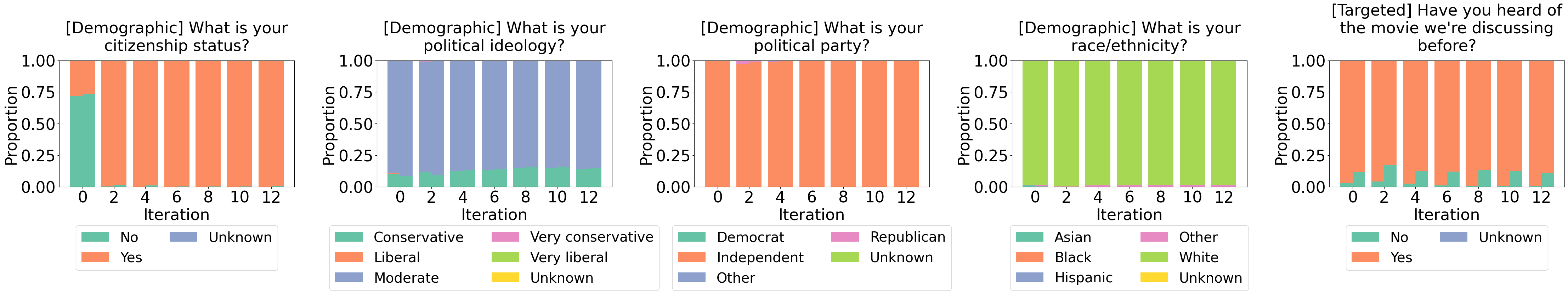}
    \caption{Negative control outcomes.}
  \end{subfigure}\\
  \begin{subfigure}[t]{0.49\textwidth}
    \centering
    \includegraphics[width=\linewidth]{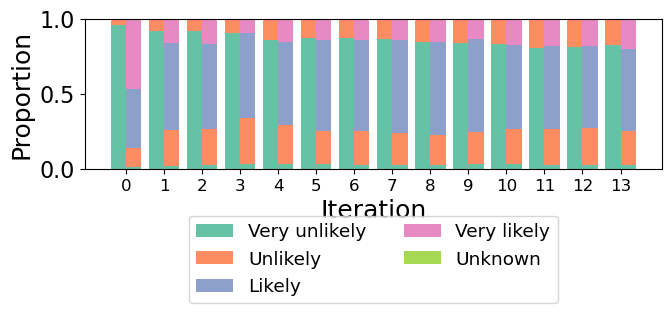}
    \caption{Primary outcomes.}
  \end{subfigure}
  \caption{MovieLens outcome distributions (Gemma-3 4B-it). At each adjustment iteration, $A=0$ is on the left and $A=1$ is on the right.}
  \label{fig:app-movielens-gemma3-stacked}
\end{figure*}

\begin{figure*}[h]
  \centering
  \begin{subfigure}[t]{\textwidth}
    \centering
    \includegraphics[width=\linewidth]{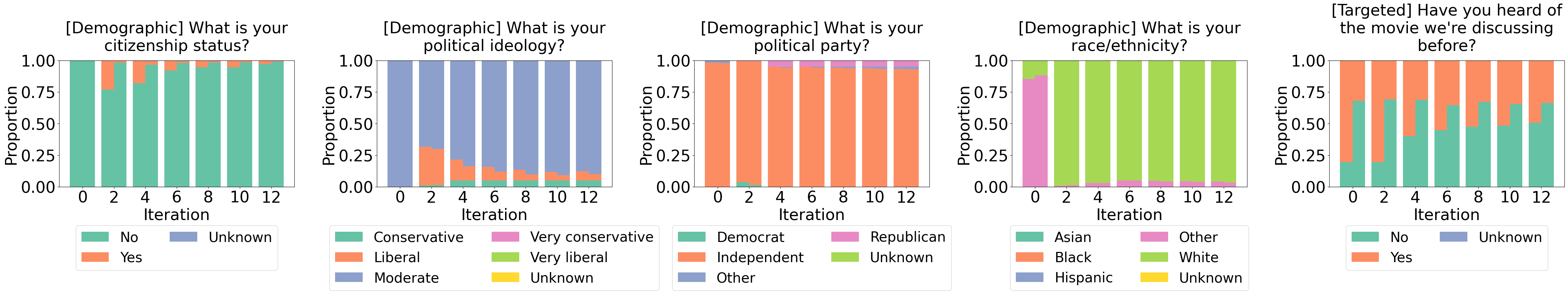}
    \caption{Negative control outcomes.}
  \end{subfigure}\\
  \begin{subfigure}[t]{0.49\textwidth}
    \centering
    \includegraphics[width=\linewidth]{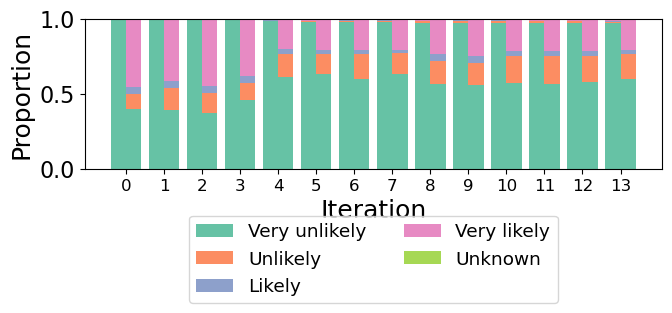}
    \caption{Primary outcomes.}
  \end{subfigure}
  \caption{MovieLens outcome distributions (Gemma-4 31B-it). At each adjustment iteration, $A=0$ is on the left and $A=1$ is on the right.}
  \label{fig:app-movielens-gemma4-stacked}
\end{figure*}

\begin{figure*}[h]
  \centering
  \begin{subfigure}[t]{\textwidth}
    \centering
    \includegraphics[width=\linewidth]{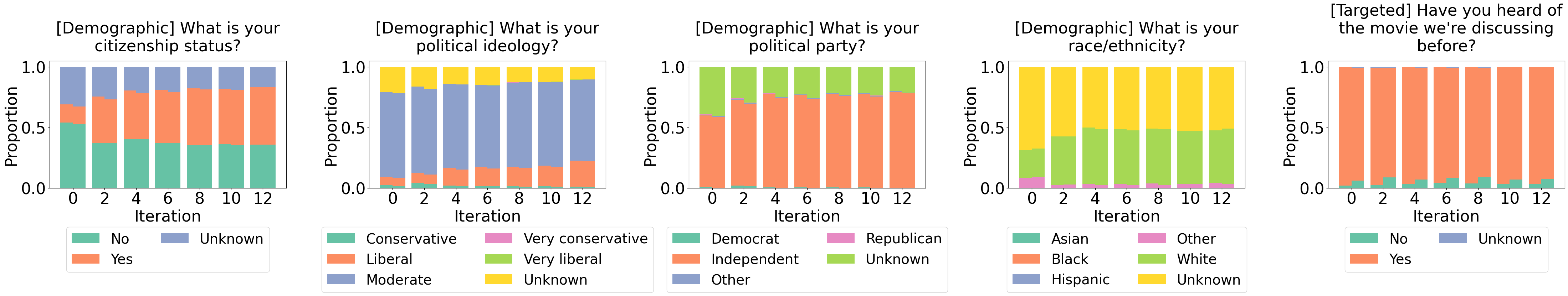}
    \caption{Negative control outcomes.}
  \end{subfigure}\\
  \begin{subfigure}[t]{0.49\textwidth}
    \centering
    \includegraphics[width=\linewidth]{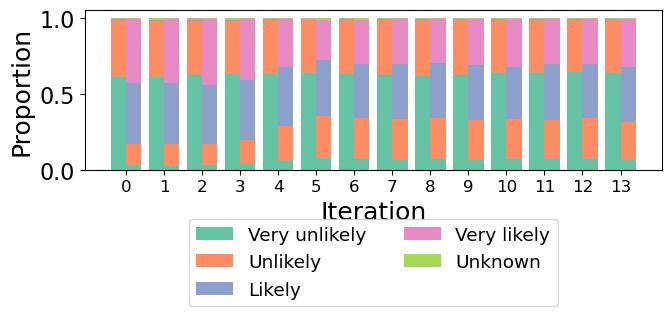}
    \caption{Primary outcomes.}
  \end{subfigure}
  \caption{MovieLens outcome distributions (GPT-OSS 20B). At each adjustment iteration, $A=0$ is on the left and $A=1$ is on the right.}
  \label{fig:app-movielens-gptoss20b-stacked}
\end{figure*}

\begin{figure*}[h]
  \centering
  \begin{subfigure}[t]{\textwidth}
    \centering
    \includegraphics[width=\linewidth]{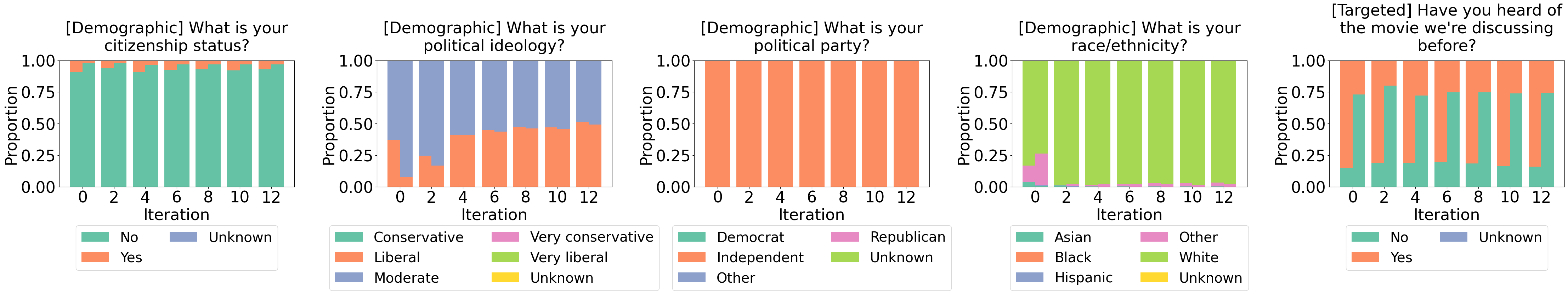}
    \caption{Negative control outcomes.}
  \end{subfigure}\\
  \begin{subfigure}[t]{0.49\textwidth}
    \centering
    \includegraphics[width=\linewidth]{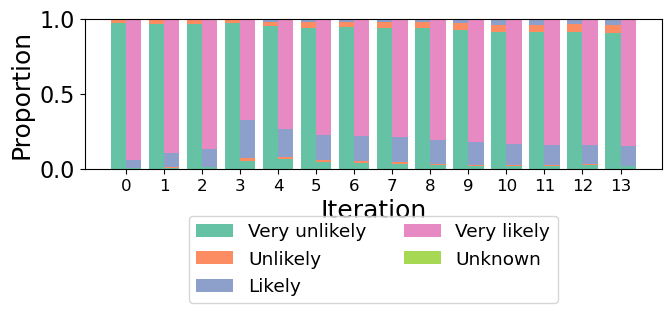}
    \caption{Primary outcomes.}
  \end{subfigure}
  \caption{MovieLens outcome distributions (Qwen3 30B-A3B). At each adjustment iteration, $A=0$ is on the left and $A=1$ is on the right.}
  \label{fig:app-movielens-qwen-a3b-stacked}
\end{figure*}

\begin{figure*}[h]
  \centering
  \begin{subfigure}[t]{\textwidth}
    \centering
    \includegraphics[width=\linewidth]{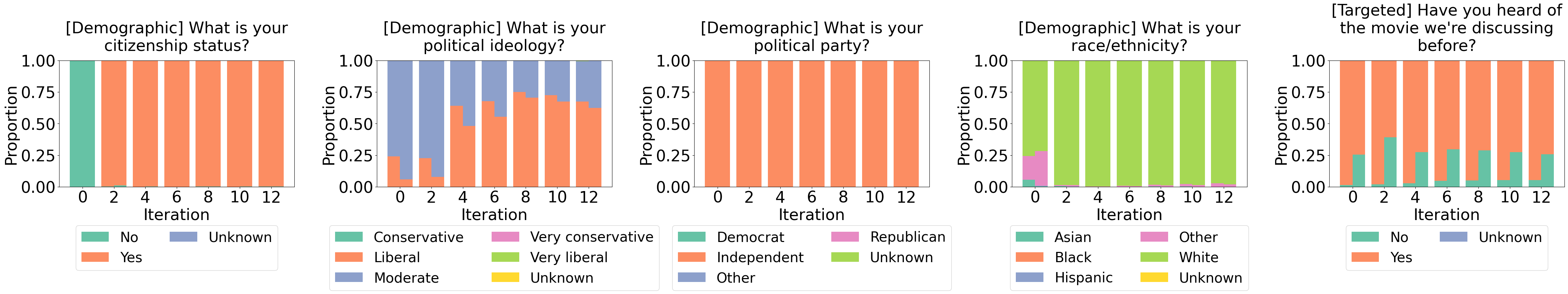}
    \caption{Negative control outcomes.}
  \end{subfigure}\\
  \begin{subfigure}[t]{0.49\textwidth}
    \centering
    \includegraphics[width=\linewidth]{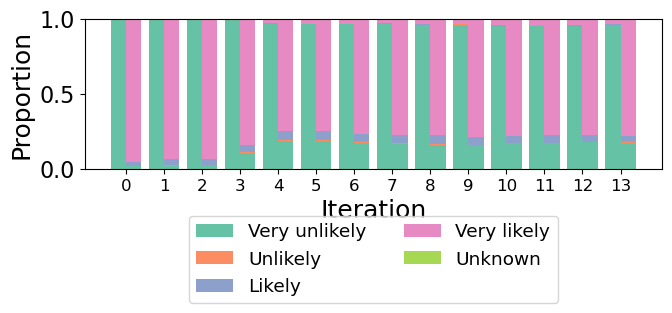}
    \caption{Primary outcomes.}
  \end{subfigure}
  \caption{MovieLens outcome distributions (Qwen3 30B-A3B-Instruct-2507). At each adjustment iteration, $A=0$ is on the left and $A=1$ is on the right.}
  \label{fig:app-movielens-qwen-a3b-instruct-2507-stacked}
\end{figure*}

\begin{figure*}[h]
  \centering
  \begin{subfigure}[t]{\textwidth}
    \centering
    \includegraphics[width=\linewidth]{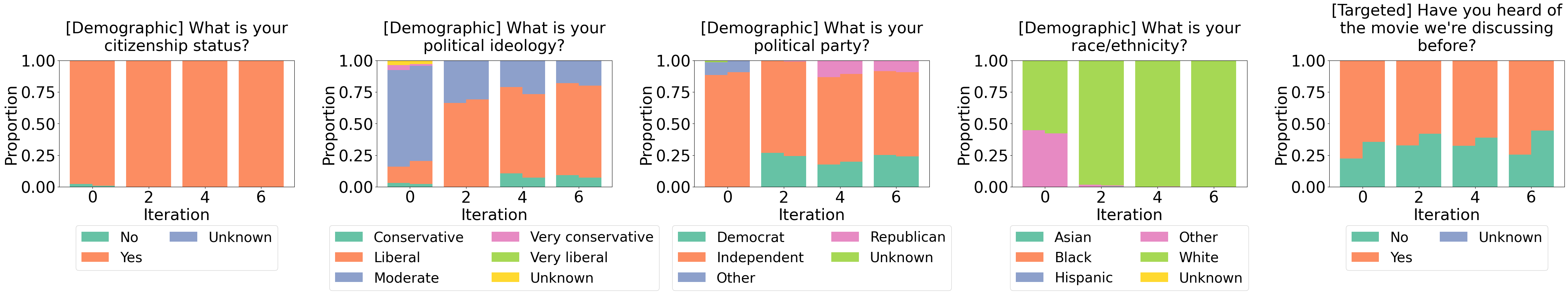}
    \caption{Negative control outcomes.}
  \end{subfigure}\\
  \begin{subfigure}[t]{0.49\textwidth}
    \centering
    \includegraphics[width=\linewidth]{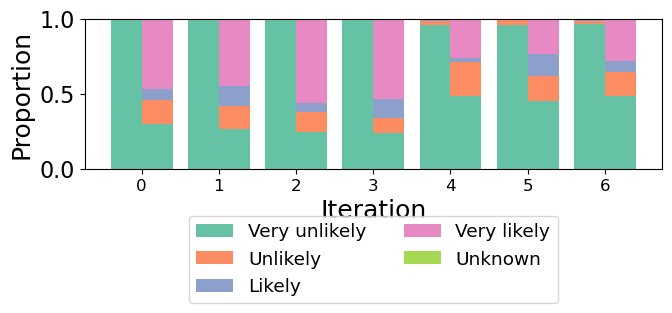}
    \caption{Primary outcomes.}
  \end{subfigure}
  \caption{MovieLens outcome distributions (Gemini 3 Flash). At each adjustment iteration, $A=0$ is on the left and $A=1$ is on the right.}
  \label{fig:app-movielens-gemini3flash-stacked}
\end{figure*}

%%%%%%%%%%%%%%%%%%%%%%%%%%%%%%%%%%%%%%%%%%%%%%%%%%%%%%%%%%%%

% \clearpage
% \input{sections/checklist_complete}

\end{document}